\documentclass[journal]{IEEEtran}
\usepackage[utf8]{inputenc}
\usepackage[T1]{fontenc}

\usepackage[
    maxbibnames=50,
    natbib=true,
    backend=biber,
    style=authoryear,
    doi=false,
    isbn=false,
    giveninits=true,
    citestyle=authoryear,
    uniquename=init
]{biblatex} 
\renewbibmacro{in:}{}
\DeclareFieldFormat*{title}{#1}

\DeclareNameAlias{sortname}{last-first}

\usepackage{framed}
\usepackage{siunitx}
\usepackage{amsthm}
\usepackage{amsfonts}
\usepackage{mathtools}
\usepackage[Symbolsmallscale]{upgreek}
\usepackage{xcolor}
\usepackage{floatrow}
\usepackage{makecell}
\usepackage{ifpdf}
\usepackage{array}
\usepackage[caption=false,font=footnotesize]{subfig}
\usepackage{comment}
\usepackage{bm}
\usepackage{multirow}
\usepackage{textcomp}
\usepackage{amsmath}
\usepackage{amstext}
\usepackage{graphicx}
\usepackage{color,soul}
\newcolumntype{P}[1]{>{\centering\arraybackslash}p{#1}}

\newcommand{\fullfgrref}[1]{\fgrref{#1} on page~\pageref{#1}}

\newcommand{\biggaussian}[3]{\mathcal{N}\left(#1 \Big\vert #2, #3\right)}
\newcommand{\mat}[1]{\mathbf{#1}}
\renewcommand{\vec}[1]{\mathbf{#1}}
\newcommand{\vecmu}{\boldsymbol{\upmu}}
\newcommand{\vecbeta}{\boldsymbol{\upbeta}}
\newcommand{\gaussian}[3]{\mathcal{N}\left(#1 \mid #2, #3\right)}
\newcommand{\fgrref}[1]{\figurename\ \ref{#1}}
\newcommand{\fgr}{\figurename\ }

\newcommand\norm[1]{l_2\left(#1\right)}
\usepackage{xr-hyper}
\usepackage{hyperref}
\usepackage{cleveref}

\colorlet{shadecolor}{yellow}

\begin{document}

\title{QTIP: Quick simulation-based adaptation of Traffic model per Incident Parameters}

\author{
    Inon Peled, 
    Raghuveer Kamalakar, 
    Carlos Lima Azevedo,
    Francisco C. Pereira
    \thanks{Emails: \{inonpe, climaz, camara\}@dtu.dk, raghu1112@gmail.com.}
    \thanks{All authors are with the Department of Technology, Management and Economics, Technical University of Denmark (DTU), 2800 Kgs. Lyngby, Denmark.}
    \thanks{Manuscript submitted to Journal of Simulation.}
}

\maketitle

\begin{abstract}
Current data-driven traffic prediction models are usually trained with large datasets, e.g. several months of speeds and flows. Such models provide very good fit for ordinary road conditions, but often fail just when they are most needed: when traffic suffers a sudden and significant disruption, such as a road incident. In this work, we describe QTIP: a simulation-based framework for quasi-instantaneous adaptation of prediction models upon traffic disruption. In a nutshell, QTIP performs real-time simulations of the affected road for multiple scenarios, analyzes the results, and suggests a change to an ordinary prediction model accordingly. QTIP constructs the simulated scenarios per properties of the incident, as conveyed by immediate distress signals from affected vehicles. Such real-time signals are provided by In-Vehicle Monitor Systems, which are becoming increasingly prevalent world-wide. We experiment QTIP in a case study of a Danish motorway, and the results show that QTIP can improve traffic prediction in the first critical minutes of road incidents.
\end{abstract}

\begin{IEEEkeywords}
Simulation, incidents, model adaptation, In-Vehicle Monitor Systems (IVMS), Intelligent Transportation Systems (ITS).
\end{IEEEkeywords}

\IEEEpeerreviewmaketitle

\section{Introduction} \label{sec:introduction}

Non-recurrent traffic disruptions are a major source of travel delays and air pollution in urban environments \citep{vlahogianni2010freeway, lee2012tim}. 
As urban traffic around the world increases constantly, main roads encounter more vehicle breakdowns, crashes, adverse weather, and large public events \citep{kwon2006components}. 
Consequently, a growing amount of resources is being invested world-wide in the study and treatment of traffic incidents \citep{mir2016kalman, wang2010parallel, kong2013developing, bertini2005response}.

Prediction models form a key component of traffic incident management for both short-term operations and long-term planning \citep{ben1998dynamit}.
Nevertheless, research into traffic prediction has concentrated mostly on incident-free conditions \citep{castro2009online, salamanis2017identifying}.
In addition, prediction models in practical use often rely on commonly available traffic data streams, as e.g. generated by mobile sensors and on-road cameras \citep{wu2012online}.
Alas, such models are slow to adapt to sudden traffic disruptions, during which effective incident treatment is most needed.

\subsection{The Challenge: Just-in-Time Model Adaptation} \label{sec:challenge}

Under ordinary conditions, speeds and travel times tend to follow consistent trends, hence real-time predictions can be made to often come close to actual values.
However, under non-recurrent disruptions, the accuracy of real-time predictions can deteriorate greatly, and dedicated methods are needed for increased accuracy \citep{chung2013TITS}. 
Indeed, immediate adaptation of traffic prediction models to sudden disruptions using real-time data has so far been a largely unsolved problem.
Several approaches to online model adaptation have been proposed \citep{wu2012online, castro2009online, ni2014using}, showing that model adaptation is needed to prevent significant deviation of predicted values (e.g., mean speed) from actual measurements.
Nevertheless, these existing approaches all assume a time buffer for adaptation, namely, they yield an adapted model only after collecting online traffic data for a few minutes following an incident.

Nowadays, however, more and more vehicles are being equipped with In-Vehicle Monitoring Systems (IVMS) \citep{viereckl2016connected, ecall}, which communicate real-time distress signals upon vehicle breakdown. 
IVMS thus offers a two-fold opportunity for online model adaptation: (1) immediate triggering, and (2) additional information about the particular circumstances of the incident.
While predicting the occurrence of traffic incidents remains a challenge in itself \citep{katrakazas2018TITS}, this paper provides empirical evidence for the possibility of quick model adaptation once an incident is known to have occurred.

\subsection{Our Contributions} \label{sec:contributions}

The prime contribution of this paper is QTIP: a novel framework for Quick Adaptation of Traffic Model per Incident Parameters. 
The novelty here lies in the combination of two traditionally separate approaches for traffic modeling, namely: data-driven machine learning and classic transport engineering methods.
To realize the benefits of this combination, let us now present the two approaches and their complementary aspects.

Given a modeling problem (e.g., speed prediction in this paper), \emph{data-driven machine learning} uses algorithms to automatically extract useful patterns from corresponding observation data.
The data consists of response variables (e.g., speed) and explanatory variables (e.g., time of day and weather).
The algorithms themselves may be either parametric or non-parametric, depending on whether or not they assume a particular functional relationship between response and explanatory variables.
Machine Learning algorithms thus serve as \emph{black boxes} that take structured data as input and yield trained models as output.\footnote{We are oversimplifying here for the sake of the argument, as there are also ``white-box'' machine learning approaches, such as Probabilistic Graphical Models \citep{pereira2019mobility}.}

On the other hand, \emph{classic transport engineering methods} are more principled and oriented towards incorporation of "behavior" in modeling.
For instance, analytical formulations of dynamic traffic assignment \citep{boyce2001analytical} use origin-destination matrices and network topology to assign traffic flows on the network, and micro-simulators for traffic modeling rely on detailed specification of driver behavior (e.g., safety distance, braking, lane changing, rubbernecking) and road conditions (e.g. presence of pedestrians, ratio of heavy to light vehicles).
Such methods are thus concerned with detailed specification of the problem through \emph{white boxes} which allow close access to underlying dynamics.

QTIP, the proposed traffic modeling framework, takes advantage of both these approaches in a manner that depends on road conditions.
Under normal traffic conditions, QTIP yields a purely data-driven model which is constructed from historical observations of ordinary traffic, as is common practice.
For abnormal traffic conditions, however, QTIP generates multiple simulations, which reflect the likely range of specific properties of the road incident, and uses their output as data for fitting a specialized machine learning model for the incident.
QTIP thus enables the use of powerful machine learning methods not only under normal and repetitive traffic conditions, but also in the first critical minutes of non-recurrent incidents.

In conclusion, we hereby enumerate all contributions of this paper.
\begin{enumerate}
    \item QTIP: a solution methodology for quasi-instantaneous adaptation of traffic prediction models, based on a novel combination of traditionally separate modeling approaches: "black-box" machine learning and "white-box" transport engineering methods.
    \item Empirical case study for a major motorway in Denmark measurably demonstrates:
    \begin{enumerate}
    \item The uniqueness of each incident.
    \item The degradation of prediction models under road incidents.
    \item The potential of QTIP in mitigating this degradation.
    \end{enumerate}
    \item Code for generating and analyzing simulated scenarios given incident information is shared in \hbox{\url{https://github.com/inon-peled/qtip_code_pub}}.
\end{enumerate}

\subsection{Paper Organization}

The rest of this paper is organized as follows.
Section \ref{sec:literary_review} provides a literary review of current solutions for real-time incident modeling. 
Section \ref{sec:qtip_description} then describes the QTIP framework, and section \ref{sec:case_study} defines a case study for evaluating QTIP. 
Section \ref{sec:results} provides the results of the case study, and section \ref{sec:discussion} concludes with a summary of our findings.

\section{Current Solutions, Gaps and Opportunities} \label{sec:literary_review}

In this section, we first review current solutions for modeling atypical traffic conditions, and the usefulness of simulations for studying incident conditions. 
Then, we motivate the necessity of online incident simulations for timely adaptation of data-driven prediction models. 
Finally, we describe a newly emerging source of real-time incident information, which is highly useful for just-in-time modeling.

\subsection{Current Methods for Traffic Prediction Under Atypical Conditions}

Accurate short-term traffic prediction is essential for proactive applications of Intelligent Transport Systems (ITS), such as Advanced Traveller Information Systems, Dynamic Route Guidance, and Traffic Control \citep{guo2012short}. Non-recurrent road incidents disrupt normal traffic patterns, and so increase uncertainty about the near future state of traffic, which thus becomes more challenging to predict. Nevertheless, traffic prediction literature has dealt much more with normal conditions than with incident conditions \citep{castro2009online, salamanis2017identifying}.

The few studies which do cater for both ordinary and incident conditions often use exclusively the data-driven approach \citep{zhang2011TITS}. Salamanis et al. \citep{salamanis2017identifying} analyze $10$ years of traffic flow and incident data under the assumption that incidents can be categorized into easily identifiable classes, and cluster the data accordingly. Thereafter, they fit $k$-Nearest Neighbors (kNN), Support Vector Regression (SVR), and Autoregressive Integrated Moving Average (ARIMA) models to each cluster, and conclude that $5$ minute prediction accuracy improves when selecting a best performing model per traffic in the preceding hour. In \citep{guo2010comparison, guo2012short, guo2014novel}, Guo et al. successively improve a set of tools for traffic prediction under normal and incident conditions. Their data-driven framework boosts performance through data smoothing and error feedback, and they consistently obtain that under abnormal traffic conditions, kNN-based methods outperform other prediction models, such as SVR, Artificial Neural Network (ANN), and Gaussian Processes (GP). The data-driven models by Salamanis et al. and Guo et al. are thus trained offline, and do not use real-time information from incidents.

Real-time model adaptation to abrupt changes in traffic conditions has been an active research subject in recent years. Wu et al. \citep{wu2012online} develop an Online Boosting Non-Parametric Regression (OBNR) model for transitioning between normal and incident conditions. OBNR is thus non-parametric, and relies on historical records for online adaptation. Castro-Neto et al. \citep{castro2009online} show that under atypical conditions, Online Support Vector Regression (OL-SVR) outperforms Gaussian Maximum Likelihood, Holt exponential smoothing, and ANN. The prediction quality of OL-SVR gradually improves as data from Vehicle Detector Stations accumulates over time. Ni et al. \citep{ni2014using} offer social network Twitter as a source of real-time information, which can improve prediction accuracy for traffic around large crowd events. Nevertheless, incorporating social data into real-time traffic analysis incurs some practical challenges, such as the need to collect, clean and fuse social data from multiple sources \citep{zheng2016TITS}.

\subsection{Studying the Effects of Road Incidents through Simulations} \label{sec:sim_effects}

Simulations are a widely used tool for studying both the short-term and long-term effects of road incidents \citep{fha2010}.
For short-term effects, Henchey et al. \citep{henchey2014emergency} use simulations to study emergency response, while Hawas et al. \citep{hawas2007microscopic} replicate real-world accidents to analyze car-following models.
For long-term effects, Wirtz et al. \citep{wirtz2005} study micro-simulations of incidents for proactive planning, while Baykal-Gursoy et al. \citep{baykal2006delay} use micro-simulations to compare strategies of traffic incident management, and Dia et al. \citep{dia2006traffic} simulate an Australian highway to measure the socio-economic impacts of incidents.

These former studies, as opposed to this paper, do not deal with online interfacing of simulations with machine learning for model adaptation.
Standing in contrast in this respect is a line of works by Ben Akiva et al. \citep{benakiva1994atms, ben1998dynamit, ben2002real, Ben-Akiva2010, dynamit2015} over DynaMIT: a framework for online traffic modeling through real-time simulations.
However, DynaMIT is concerned with modeling traffic conditions on a network-wide level (e.g., the overall state of congestion in a city) in the context of interactions between transport demand and supply.
Conversely, this paper focuses on incident conditions in a purposely constrained environment and assumes only limited prior knowledge of transport demand.
Our proposed solution framework is thus complementary to the overall vision of DynaMIT, and we indeed suggest to incorporate it as a component in DynaMIT (Section \ref{sec:future}).

\subsection{Motivation for Incorporating Simulations in Real-Time Incident Modeling} \label{sec:motivation}

Real-time data-driven prediction models take advantage of the relative stability of conditions over short, consecutive intervals.
For example, explanatory variables such as time-of-day, day-of-week, effects of seasonal trends, and weather typically change very little over $5$ consecutive minutes.
Hence the closer a data-driven model is to real-time resolution, the lower is its necessary complexity (e.g., non-linearity, more explanatory variables), as the effects from trends and context are already incorporated in the current time window.

In contrast, when an incident happens, the correlation structure between response and explanatory variables changes abruptly, in a manner which is unique to the incident characteristics. 
For example, when an incident occurs, the mean speed in the current time interval may significantly change its pattern of dependency on speeds in recent time intervals, in the affected link and in its neighboring links. 
As such, there are advantages to treating incidents separately from other atypical conditions, through a dedicated prediction modeling framework.

On one hand then, data-driven prediction models tend to break under the sudden change of correlation structure brought about by an incident.
On the other hand, such models could perform well if their input data adequately pertained to the given incident.
Unfortunately, historical road incident data is often deficient or altogether missing \citep{kumar2015data}.

At first sight, it may seem worthwhile to try and overcome this lack of data by pre-generating sufficiently many incident simulations offline.
However, such attempts would in fact be impractical, because every incident involves too many varying parameters (e.g., location, road attributes, weather conditions, lane occupancy levels, vehicles involved, proximity to control systems, etc.).
Consequently, only a very restricted subset of all possible combinations of incident parameters can be covered offline, and a data-driven model trained in such manner will necessarily underperform on some out-of-sample scenarios.

It follows that to successfully take advantage of data-driven methods, a framework for prediction modeling under real-time incidents must generate online data which corresponds to the unique parameters of each incident occurrence.
This motivates us to consider real-time simulations as a means of generating such data online.
For these online simulations, we next describe a useful and globally emerging source of real-time incident information.

\subsection{Real-Time Incident Data for Online Modeling}

As the vision of always-connected cars (V2X) progresses worldwide \citep{siegel2017TITS}, active In-Vehicle Monitor Systems (IVMS) are becoming increasingly prevalent nowadays \citep{viereckl2016connected, Brandl2016}. In fact, certain IVMS systems are now mandatory by law, e.g. the European Union now mandates that the eCall system be installed in every new vehicle \citep{ecall}. Real-time signals from IVMS are designed to indicate the status and location of vehicles, and in particular, the occurrence of malfunctions and crashes \citep{ivmsDigicore}.

Therefore, IVMS delivers not only immediate indication of incident occurrence, but also rich information about the particular properties of the incident. 
In the next section, we present a framework which employs IVMS signals in real-time simulations for corresponding prediction model adaptation.

\section{The QTIP Generic Framework} \label{sec:qtip_description}

\subsection{Overall Framework}  \label{sec:overall_framework}

\begin{figure}[t]
    \centering
    \includegraphics[width=\textwidth]{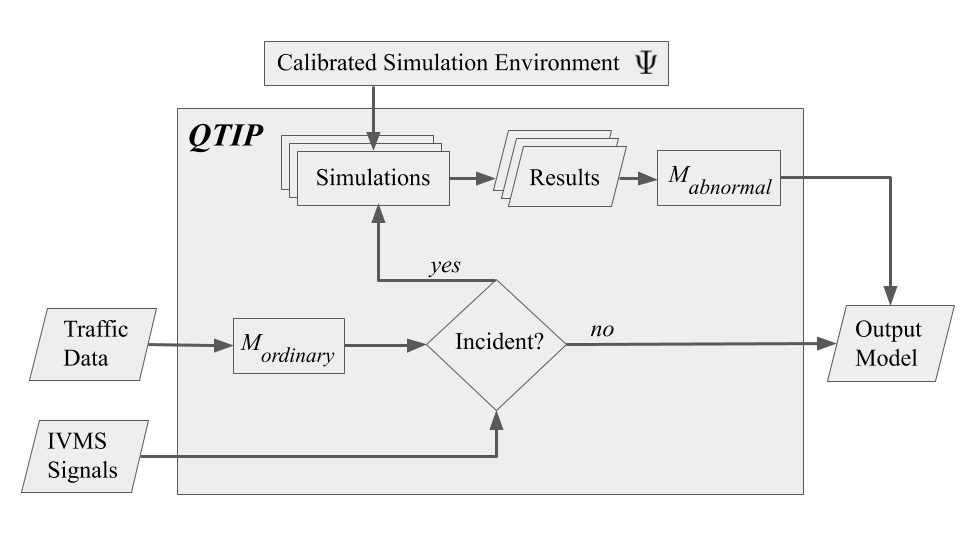}
    \caption{QTIP Framework.}
    \label{fig:qtip_operation_diagram}
\end{figure}

In this section, we present the framework of QTIP, and illustrate how it differs from current solutions. In the remaining sections thereafter, we evaluate several different instances of this framework against an experimental case study, and show how QTIP can address the first critical minutes of road incidents.

\fgr \ref{fig:qtip_operation_diagram} summarizes the main components of our QTIP framework. 
The input to QTIP consists of common data about traffic -- e.g. from road sensors, mobile sensors, and weather stations -- and incident-specific information from IVMS. 
QTIP uses both data streams to output a traffic prediction model, as follows.

On one hand, when no incident is known to have occurred, QTIP directly yields model $M_{ordinary}$, which is fit for incident-free conditions.
For example, $M_{ordinary}$ can be any desired data-driven model, pre-trained on historical records and consistently updated on recent traffic data.

On the other hand, upon receiving IVMS signals from vehicles involved in a road incident, QTIP yields an adapted traffic prediction model $M_{abnormal}$ by executing multiple simulations.
Let us now elaborate on the purpose of these incident simulations and the manner in which they are implemented.

The purpose of the incident simulations is to cover a range of \emph{unobserved} explanatory variables that determine how severely the incident affects its surroundings.
In this paper, we use two such variables as example: level of road usage (i.e., "traffic demand") at the moment of the incident, and the precise position of the incident.
Each executed simulation thus pertains to a different combination of possible values for the unobserved variables, while also accounting for the \emph{observed} information in the IVMS distress signals, e.g., time of occurrence, number of signals, and general location on the road network.

The simulations are then implemented through two main steps.
First, a simulated environment of the affected road (e.g., the motorway in our case study) is constructed and calibrated to resemble its real-world structure. 
This step requires a dedicated solution component, which we denote in \figurename {\ } {\ref{fig:qtip_operation_diagram}} as $\Psi$, and which we purposely leave out to future work.

Admittedly, we do not intend to offer here a complete and operational solution, but rather provide a theoretical study of challenges and benefits in extracting value from real-time signals with limited incident information. 
This study does show that even partial incident information -- e.g., one that lacks data about current traffic demand -- can still be useful for noticeably improving traffic prediction quality.
We also note that pre-calibrated simulation environments can be prepared in advance for roads that are known to be incident-prone, so that QTIP is ready to simulate real-time incidents per their unique characteristics.
In fact, the QTIP case study in this paper uses such a pre-calibrated simulation environment for an incident-prone motorway in Denmark.

Once the simulated environment (namely, the affected road) is constructed and calibrated, the second step is to use it for executing the desired simulations.
To this end, we use PTV VISSIM as the underlying micro-simulation engine and utilize its Component Object Model (COM)-based Application Programming Interface (API).
Through this API, we bootstrap each simulation in real-time per the corresponding variable values -- both observed and unobserved -- and run all simulations in parallel.
QTIP then uses the simulation results to fit and output $M_{abnormal}$, the adapted prediction model. 

\subsection{Advantages of QTIP over Existing Solutions} \label{sec:advantages}

Let us now highlight several desirable properties of the QTIP framework, which current solutions lack to some extent, as reviewed in Section \ref{sec:literary_review}. First and foremost, QTIP is designed to readily take advantage of information from the incident itself, as the change in correlation structure between response and explanatory variables is unique to each incident. And so, whereas the prediction quality of e.g. OL-SVR \citep{castro2009online} gradually improves over time, QTIP yields a completely adapted model shortly after incident parameters are known. As we show in section \ref{sec:case_study}, a few incident parameters could indeed be enough for QTIP to yield an effective new model.

Second, QTIP is agnostic to the specific form of models $M_{ordinary}$ and $M_{abnormal}$.
These models can thus be chosen freely, e.g. as parametric and interpretable models, as we further elaborate in Section {\ref{sec:model-selection}}.
Hence whereas OBNR \citep{wu2012online} is non-parametric, QTIP easily allows insights into how a prediction model changes when adapted to different traffic conditions. Furthermore, OBNR relies on historical records for online adaptation, whereas QTIP does not require past examples of incidents to yield an adapted model.

In fact, QTIP is also agnostic to the specifics of its input traffic data, which can thus consist of both sensor readings and relevant feeds from social networks, as suggested by Ni et al. \citep{ni2014using}. Nevertheless, while QTIP welcomes such contextual information, its immediate response relies only on signals which originate directly from the road incident.

Overall then, none of the current solutions relies primarily and systematically on real-time information about the incident itself. It is thus questionable whether discrete data-driven approaches, such as the clustering method of Salamanis et al. \citep{salamanis2017identifying}, can solve the problem of abrupt changes to traffic correlation structure. Furthermore, as QTIP employs real-time simulations, it can be used as a component within other systems for real-time Dynamic Traffic Assignment, such as DynaMIT \citep{dynamit2015}.

\subsection{Model Selection} \label{sec:model-selection}

As mentioned above, the traffic models in QTIP can be freely chosen, hence this work focuses on proof-of-concept of the working principles of QTIP.
We also note that, as observed in \citep{tune2016comparison}, the range of traffic models is already too large to examine in detail here.
Consequently, we refrain from making particular recommendations on specific model types for modeling ordinary and abnormal traffic conditions.
However, for completeness of description, we now provide several effective guidelines for model selection; a broader discussion of model selection is available in \citep{claeskens2008model}.

Model selection entails a choice between several \emph{model classes}, such as:
\begin{itemize}
    \item \emph{Parametric} vs. \emph{non-parametric} modeling \citep{smith2002comparison}, namely, particular functional form vs. flexible form.
    \item \emph{Simple} vs. \emph{complex} models \citep{hoogendoorn2001state}, e.g., as measured by number of parameters: a sufficient number is needed to properly generalize to unseen data, while an improper number might lead to under-fitting or over-fitting.
    \item \emph{Interpretable} vs. \emph{opaque} models \citep{wang2020interpretable}: the former may allow better explanation of the captured patterns, but restricts the range of applicable models; the latter allows less in-depth insights, but admits a wider selection of models.
    \item \emph{Standalone} models vs. \emph{ensemble} methods \citep{li2014multimodel}: the latter may improve accuracy by combining the strengths of multiple models, but also requires further weight tuning and might introduce unnecessary complexity.
\end{itemize}
Accordingly, we later experiment with models from several different classes.
Moreover, as the method of \emph{estimating model parameters} is also a matter of choice, we experiment with both Maximum Likelihood Estimation (Section \ref{sec:experiments}) and Bayesian Inference (Section \ref{sec:transfer-learning}).

Once several models are selected for comparison, they can be compared using measures of \emph{goodness of fit} \citep{dagostino1986goodness}, including:
\begin{itemize}
    \item \emph{Prediction quality} analysis, e.g., via Mean Error measures or Coefficient of Determination.
    \item \emph{Statistical tests} of similarity between fit and expected distribution, e.g., via Analysis of Variance or hypothesis tests, such as the Kolmogorov-Smirnov Test and Pearson's Chi-Squared Test.
\end{itemize}
For example, in this work, we compare models based on minimization of MSD, MAE and RMSE (Section \ref{sec:degradation}) between model predictions and test observations.

In addition, there exist various \emph{Information Criteria} (IC) for model selection, based on theoretically optimal tradeoffs between model simplicity and fit accuracy \citep{tune2016comparison}.
Different IC employ different assumptions about the modeled data, and popular IC examples include the Akaike Information Criterion, Bayesian Information Criterion, and Minimum Description Length.
In this work, measures of prediction quality suffice for choosing between the models, hence we do not employ additional IC.

In summation, for traffic modeling in QTIP too, we advise to follow the systematic guidelines of model selection as above.
Furthermore, we next describe a proof-of-concept case study for evaluating QTIP, where we employ various models of speed prediction in a motorway.
The case study thus constitutes an example of using QTIP itself to select between models through simulations, particularly for accident-prone roads.
This simulation-based manner of model selection can be augmented by incorporating additional historical data about other roads in the traffic network, including records of past accidents.

We note that while speed prediction alone does not suffice for incident treatment by traffic practitioners, the principles of simulation-based model adaptation in the case study extend to other traffic models -- such as traffic flow, trip delays, public transport disruptions, and risk of secondary incidents -- which together are highly useful for incident management.
The speed prediction models that we study can also provide road users with real-time information about the occurrence and effect of incidents, which is useful for route planning and estimation of trip time.

\section{Case Study for Evaluating QTIP} \label{sec:case_study}

In this section, we present experiments to evaluate the performance and capabilities of a specific application of QTIP. As a case study, we use the Hiller{\o}d Motorway, a highly utilized and often congested highway in Denmark, and compare the performance of a model adapted by QTIP vs. a non-adapted model in incident conditions.

\subsection{Overall Design of Experiments} \label{subsec:experiments}

To evaluate the QTIP framework, we first need to decide which type of models $M_{ordinary}$ and $M_{abnormal}$ are used (see \fgr \ref{fig:qtip_operation_diagram}). We chose to evaluate three types of models separately: Linear Regression (LR), Deep Neural Network (DNN), and Gaussian Processes (GP). Whereas LR is linear and parametric, DNN and GP are non-linear and non-parametric.  For each prediction model type, we pre-train $M_{ordinary}$ on a set of incident-free simulations, which we describe in the next section. In this manner, we generate different demand and road conditions in the training set, and can test how the proposed model types differ in performance.

\begin{figure}[t]
    \centering
    \includegraphics[scale=0.4]{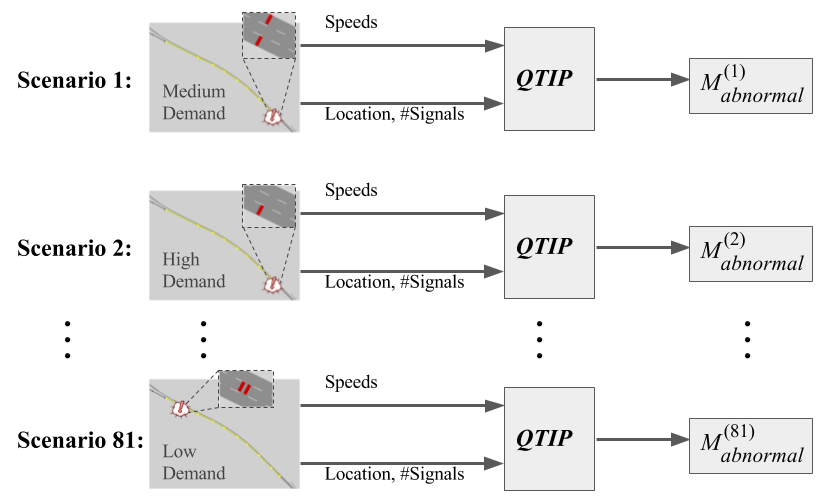}
    \caption{In our experiments of incident conditions, we run QTIP to obtain $M_{abnormal}$ for different incident scenarios independently.}
    \label{fig:qtip_usage}
\end{figure}

Finally, we compare between the simple, non-adaptive model $M_{ordinary}$ and QTIP on several incident scenario experiments (see \fgr \ref{fig:qtip_usage}). In the present work, all considered incident scenarios are represented as blocks on a given road link. Each scenario differs in: (1) the location of road blocks in the link; (2) the location of road blocks in lanes; (3) the number of road blocks; and (4) the demand in terms of traffic volume.

\begin{figure}[t]
    \centering
    \includegraphics[scale=0.6]{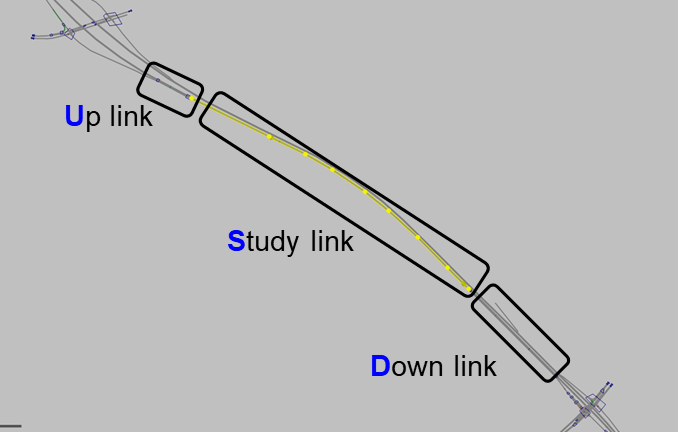}
    \caption{The Study Link $S$ where we simulate incidents on the Hiller{\o}d Motorway, its uplink $U$, and downlink $D$. Traffic on these links flows uni-directionally north-to-south.}
    \label{fig:links}
\end{figure}

For each incident scenario, we generate multiple replications, independently and with stochastic perturbations. Each replication serves as a single, independent experiment. To reflect real-time operation, QTIP receives in each experiment the simulated speed measurements, averaged in $1\ min$ intervals, in three links: the link where road blocks would appear, its uplink, and its downlink, as in \fgr \ref{fig:links}.

Before the road blocks manifest, QTIP predicts future mean speeds through the already pre-trained $M_{ordinary}$, i.e. without adaptation. When the road blocks appear, QTIP receives notification of their location, similarly to IVMS distress signals. QTIP then quickly creates model $M_{abnormal}$ using "on-the-fly" simulations, and uses it for subsequent predictions.

To clarify, note that each experiment involves two different types of simulations. One is the simulation that emulates the ground truth, which provides QTIP with input of actual traffic. The other simulations are those which QTIP executes internally "on-the-fly" to construct $M_{abnormal}$. QTIP uses such internal simulations to consider "what-if" values of unknown incident parameters, which the distress signal does not indicate. For example, the distress signal does not carry the current demand level, hence while the actual demand may be "medium", QTIP internally considers all of "low", "medium", and "high" as possible demand levels in predictions. To keep this separation clear, we shall refer to the two types of simulations as "ground-truth simulations" and "what-if simulations", respectively.

\subsection{The Road Network}

\begin{figure}[t]
    \centering
    \includegraphics[width=\textwidth]{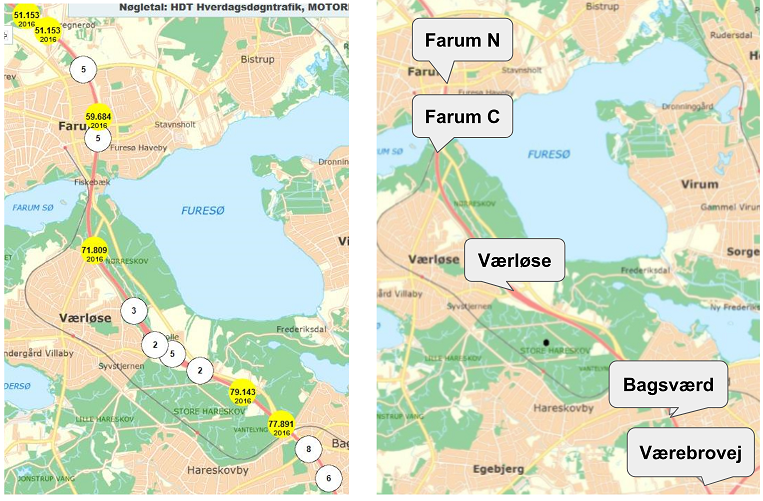}
    \caption{The Hiller{\o}d Motorway: data collection points as white circles (left), and locations of entries and exits (right).}
    \label{fig:hillerod}
\end{figure}

The Hiller{\o}d motorway is located in Sj{\ae}lland region in the Greater Copenhagen Area (\fgr \ref{fig:hillerod}) and is known for recurrent congestion and significant impacts of incidents on traffic conditions. $604$ accidents were recorded in the past $5$ years (2012-17) in the total $35.9\ km$ of its length. Furthermore, several towns and cities served by this motorway \textendash{} Farum, V{\ae}rl{\o}se and Gladsaxe \textendash{} have planned to complete several developments by 2020 \citep{trm2016}, which will increase significantly the traffic on the corridor.

For this work, we focus on the stretch between Farum N and V{\ae}rebrovej in the North/South direction, as in \fgr \ref{fig:hillerod}. The test network is in total $11\ km$ long and consists of signalized ramps, five interchanges and a two-lane carriage. $36.5$\% of all observed accidents (i.e. $220$ records) in the past $5$ years were in this stretch, $5.3$\% of which resulted in injuries and/or fatalities.  The recurrent congestion and disruptions already motivated the implementation of a $3\ km$ Hard Shoulder Running stretch, active during the morning peak hours \citep{DRD2016}. 

\subsection{Data for Calibration} \label{sec:data}

To calibrate the simulated environment, we used three types of data, all from the Danish Road Directorate\footnote{Mastra and Hastrid databases, \href{http://www.vejdirektoratet.dk/}{http://www.vejdirektoratet.dk/}}: flows, average speed, and travel times, by link and in time intervals of $15$ minutes (see Table \ref{table_vehicle_flows_on_field} for vehicle volumes aggregated by hour). The data originates from eight online data collection points along the motorway, shown as white circles in \fgr \ref{fig:hillerod}.

For this paper demonstration, it sufficed to gather a limited data set of field measurements. As a calibration set, we used measurements from the 7-8 AM period of the weekdays of June 20 to 24, 2016. As a separate validation set, we used the preceding week, namely June 13 to 17, 2016. As demand input for the simulator, we constructed two Origin-Destination (OD) matrices: one for light vehicles, and one for heavy vehicles. We also conducted on-field data collection to measure some network attributes, such as speed limits and signal timings.

\begin{table}[t]
    \caption{Vehicle counts, Monday 2016-Jun-20 7-8 AM}
    \label{table_vehicle_flows_on_field}
    \centering
    \begin{tabular}{l|c|c}
    \Xhline{2\arrayrulewidth}
    \bfseries Location & \bfseries Entrance & \bfseries Exit \\
    \hline 
    Farum N & $3171$ & Network start point \\
    Farum C & $544$ & No exit point \\
    V{\ae}rl{\o}se & $937$ & $100$ \\
    Bagsv{\ae}rd & $852$ & $544$ \\
    V{\ae}rebrovej & $351$ & $312$ \\
    \Xhline{2\arrayrulewidth}
    \end{tabular}
\end{table}

\subsection{Construction and Calibration of Simulated Environment}

The simulator we use in this study is PTV VISSIM\footnote{\href{http://vision-traffic.ptvgroup.com/en-us/products/ptv-vissim/}{http://vision-traffic.ptvgroup.com/en-us/products/ptv-vissim/}}. 
VISSIM is widely applied in practice for modeling of transportation systems, reproduction of freeway driving conditions, analysis of traffic operations, and studying of incident conditions \citep{gomes2004calibration, katrakazas2018TITS}.

For constructing the simulated network, we used OpenStreetMap\footnote{\href{https://www.openstreetmap.org/}{https://www.openstreetmap.org/}} and Google Earth\footnote{\href{https://www.google.com/earth/}{https://www.google.com/earth/}} along with the above-mentioned field observations. For creating the signal systems which control the inflow of the network, we used VISVAP, VISSIM's add-on for traffic signal controls and traffic management systems
\footnote{\href{http://www.traffic-inside.com/tag/visvap-en/}{http://www.traffic-inside.com/tag/visvap-en/}}. As no control pre-set configuration was available, we performed manual tuning of fixed signaling cycles to match both the field observations and the recorded flow data.

For car-following behavior, we used the Wiedemann 99 model as in \citep{aghabayk2013novel}. This is controlled in VISSIM through parameters $CC0 \dots CC9$, together with two look ahead/back distribution parameters \citep{vissimmanual}. For lane change behavior, we considered seven parameters, which represent acceleration distribution during lane changing manoeuvres, how far in advance each driver can anticipate the next exit/weaving/lane drop, and how aggressively that driver changes lanes to reach there. \citep{gomes2004calibration}.

We calibrated the simulated environment offline through a manual iterative process, as per \citep{park2006}, while following the Danish Guidelines \citep{drd2010} for model evaluation. Because microscopic traffic simulators are characterized by a large number of parameters (usually to represent different driving behaviors), we carried out a sensitivity analysis to select a subset of parameters to calibrate, following the method in \citep{manjunatha2013methodology}, which VISSIM also uses for similar performance measures (delays, flows and speeds). Note that in our simulated road network, there is essentially only one route from each origin to destination, thus relaxing the impact of route choice parameters.

For each iteration in the calibration process, we ran enough replications until reaching a level of confidence for the average travel times, mean speeds, and vehicle flows at the data collection points \citep{hollander2008principles}. As Measures of Effectiveness (MOEs) for quality of calibration, we used Rooted Mean Squared Normalized Error ($RMSNE$) for speeds and travel times, and Geoffrey's E. Havers' value ($GEH$) for flows \citep{hollander2008principles}:
\begin{align} 
RMSNE &= 
\left( e_1/o_1 + \cdots + e_n/o_n\right)^{0.5} / n^{0.5}\\ 
GEH &= 
\left( 2 e_1^2/p_1 + \cdots + 2 e_n^2/p_n \right)^{0.5}
\end{align}
where for all $i=1..n$: $e_i = m_i - o_i$ and $p_i=m_i + o_i$, such that $o_i$ is the value actually observed and $m_i$ is the corresponding value in simulation. 
As described previously in Section \ref{sec:data}, we used a separate set of data measurements for validation.

From the sensitivity analysis, we identified $9$ parameters which had the most impact on the Measures of Effectiveness. We have calibrated these parameters as summarized in Table \ref{tab:influential_parameters}.

\begin{table*}[t]
    \centering
    \caption{Final Calibrated Parameters}
    \label{tab:influential_parameters}
    \begin{tabular}{l|c|c|c}
    \Xhline{2\arrayrulewidth}
    \bfseries Parameter & \bfseries Unit & \bfseries Calibration Range & \bfseries Calibrated Value \\
    \hline
    Desired Speed Distribution & $km/h$ & $50 \ldots 110$ & $80-110$\\
    Reduced Speed Areas & $km/h$ & $20 \ldots 30$ & $20$\\
    Emergency Stop Distance (LC) & $m$ & $5 \ldots 50$ & Varies by Link Connector \\
    Lane Change Distance for Weaving Areas (LC)& $m$ & $200 \ldots 500$ & Varies by Link Connector \\
    Maximum Deceleration for Breaking & $m/s^2$ & $3 \ldots 8$ & $3.0$\\
    Waiting Time Before Diffusion & $s$ & $10 \ldots 60$ & $60$ \\
    Standstill Distance (CC0) & $m$ & $1.5, 2.0, 2.5$ & $3$ \\
    Headway Time (CC1) & $s$ & $0.9, 1.0, 1.1$ & 1\\
    Safety Distance Reduction Factor & & $0.1 \ldots 0.6$ & $0.1$\\
    \Xhline{2\arrayrulewidth}
    \end{tabular}
\end{table*}

\begin{figure}[t]
    {\centering
    \sidesubfloat[][]{
        \includegraphics[width=0.95\textwidth]{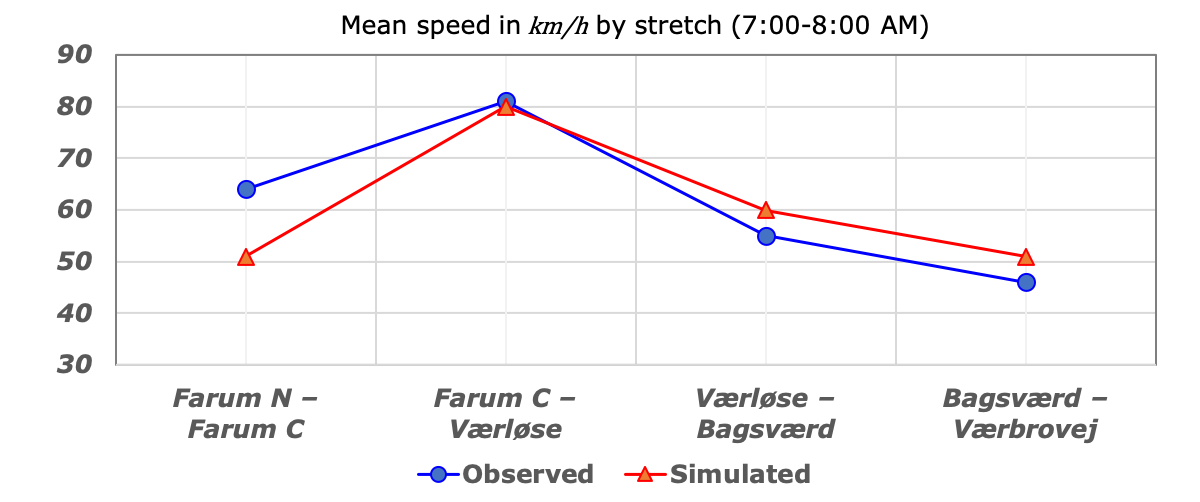}
        \label{fig:5a}
    }
    \hfil
    \sidesubfloat[][]{
        \includegraphics[width=0.94\textwidth]{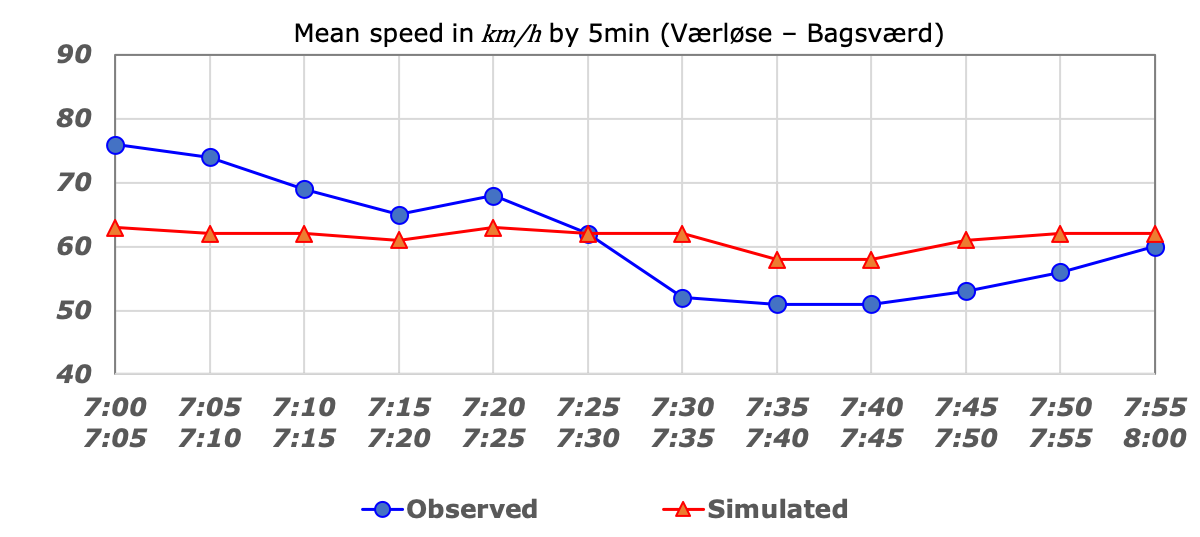}
        \label{fig:5b}
    }
    \hfil
    \sidesubfloat[][]{
        \includegraphics[width=0.95\textwidth]{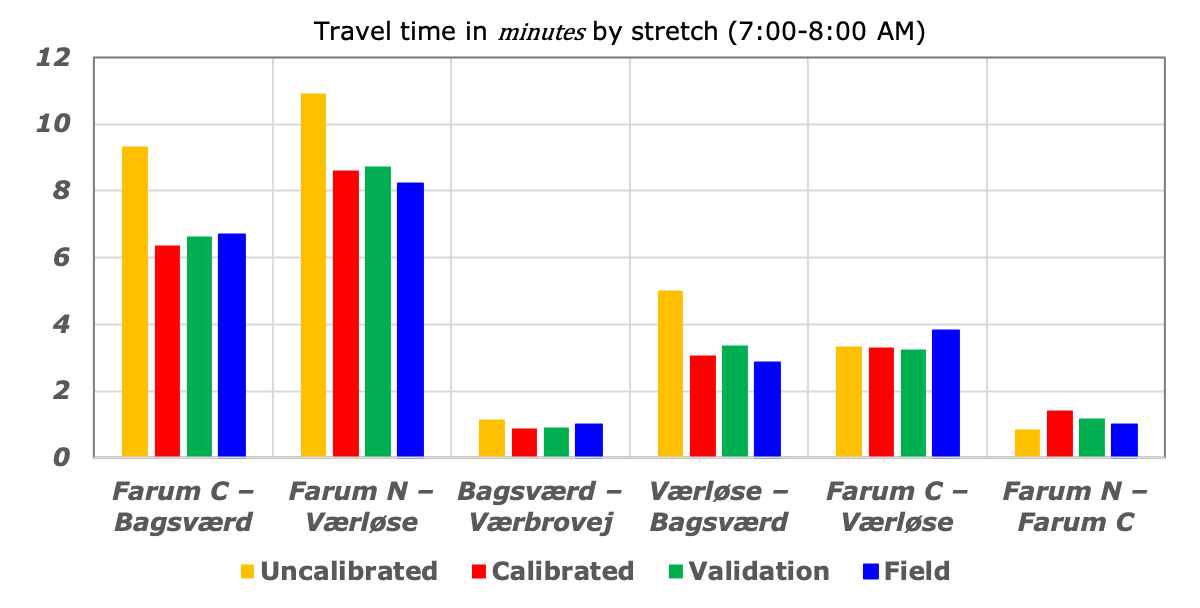}
        \label{fig:5c}
    } 
    \par}
    \caption{Results of calibration: speeds (\ref{fig:5a}, \ref{fig:5b}) and travel times (\ref{fig:5c}). In \ref{fig:5c}, "calibration" and "validation" correspond to the data sets defined in Section \ref{sec:data}.}
    \label{fig:calibrationresults}
\end{figure}

Per \cite{fdot2014}, a traffic model is acceptably calibrated when the MOEs yield $RMSNE < 0.15$ and $GEH < 5$. Eventually, we have obtained the following measurements of calibration quality:
\begin{itemize}
    \item $RMSNE = 0.12$ for speeds and for travel times.
    \item $GEH < 2$ for traffic counts.
    \item Less than $1\ min$ difference between simulated and observed mean travel times for each stretch of road, as shown in \fgr \ref{fig:5c}. In particular, the end-to-end mean travel time that the calibrated model attains is $8.23\ min$, which is close to $8.6\ min$ actually measured on-field.
    \item Stretch-to-stretch mean speeds close to measured speeds, as reflected in \fgr \ref{fig:5a} and \fgr \ref{fig:5b}.
\end{itemize}

Finally, it is worth noting that several advanced offline calibration approaches have been proposed in the literature, tackling the complexities of large scale calibration \citep{zhang2017efficient} and the high dimensionality of both input and output performance measures \citep{ciuffo2014sensitivity}. Yet, for the purpose of the case study at stake, the method we used above provided satisfactory performance.

\subsection{Experiments} \label{sec:experiments}

Having calibrated the simulated environment, we are now ready to execute experiments. Recall that we independently experiment with three different instances of QTIP: LR, DNN, and GP.

\subsubsection{Details Common to All Experiments} \label{sec:common}
Each QTIP experiment begins with constructing $M_{ordinary}$. To this end, we first run $150$ incident-free simulations -- $50$ for each demand level: low, medium, and high -- and fit $M_{ordinary}$ on the output of these simulations. Next, we proceed to run incident scenarios. In each scenario, an obstacle appears at 7:10 AM on the study link $S$, and disappears at 7:40 AM. This affects traffic not only in $S$, but also in its uplink $U$ and downlink $D$, as shown in \fgr \ref{fig:links}.

\begin{figure}[t]
    \includegraphics[width=\textwidth]{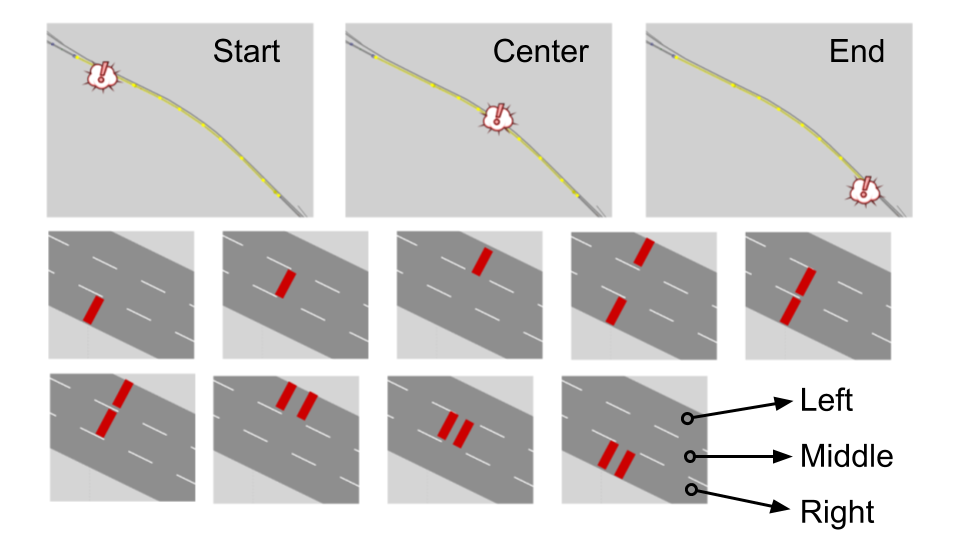}
    \caption{In our experiments, road blocks appear at one of three locations on the Study Link, and occupy one or two lanes.}
    \label{fig:location_and_lanes}
\end{figure}

\fgr \ref{fig:location_and_lanes} illustrates the various options for incident positions in our experiments, as following. Each incident scenario involves either $1$ or $2$ road blocks in one of three \emph{locations} on $S$: Start, Center, or End. Each road block is positioned on one of the three \emph{lanes} of $S$: Left, Middle, or Right. Two simultaneous road blocks occupy either the same lane, one right after the other (spaced by $10m$), or different lanes, in which case they are located next to each other. In addition, each scenario involves one of three demand levels: low, medium, or high. Hence in total, the number of distinct incident scenarios is
\begin{equation}
    81 = 3 \times 3 \times \left(B(3,1) + B(3,1) + B(3,2)\right)
    \,,
    \label{eq:81}
\end{equation}
where $B(a, b) = a!/(b!(a-b)!)$ is the binomial coefficient, and the terms on the right-hand side correspond respectively to the number of options for: demand, location, single blocked lane, two blocks in same lane, and two blocks in different lanes.


For each of these $81$ incident scenarios, we generate in VISSIM $5$ ground-truth simulations, replicated with perturbations, as we soon explain. Finally, we test QTIP independently on each ground-truth simulation in two different modes: one where the distress signals carry high precision location, and one where location precision is low. QTIP knows which specific lanes are blocked only in high precision mode, in which case it could possibly yield more accurate predictions.

When QTIP receives notification of an incident, it generates $100$ what-if simulations, and uses their results to fit $M_{abnormal}$ as a piece-wise prediction model. The first piece pertains to the first $6$ critical minutes immediately after the appearance of road blocks, and the second piece pertains to the time until the incident is cleared. $M_{abnormal}$ also takes into account the number of minutes which have passed since incident occurrence, as variable $T_{accident}$.

To recap so far, three sets of simulations are involved: incident-free (for $M_{ordinary}$ training), incident ground-truth (used as benchmark "real" measurements), and incident what-if (created by QTIP to train $M_{abnormal}$). To account for the stochastic nature of traffic, we apply perturbations to each simulation as follows: before running any simulation, we perturb its input OD matrix $A$ independently as $\tilde{A}$, so that $\tilde{A}_{ij}=\tilde{c}_{ij} A_{ij}$, where $\tilde{c}_{ij} \sim \mathcal{N} \left( 1, 0.2\right)$. To simulate different demand levels, we also scale $\hat{A} = \hat{c} \tilde{A}$, so that $\hat{c}=0.7, 1.0, 1.3$ for low, medium, and high demand, respectively. The simulation then runs with $\hat{A}$ as its OD matrix. Note that medium demand corresponds to the same demand as we have measured on-field earlier, with an added stochastic component.

All simulations yield as output $1\ min$ mean speeds in links $S, D, U$ (\fgrref{fig:links}).
Models $M_{ordinary}$ and $M_{abnormal}$ receive this output in the form of vectors, as following. 
At any time point, let $S_k, U_k, D_k$ denote the $1\ min$ mean speed $k$ minutes earlier in links $S, U, D$, respectively. For each of $\text{6:50}, \text{6:51}, \dots$, the corresponding vector is $\left[ S_5, S_6, U_5, U_6, D_5, D_6 \right]$, and $S_0$ is the response variable. That is, to predict the $1\ min$ mean speed on $S$, the models use the $1\ min$ mean speeds $5$ and $6$ minutes beforehand on links $S, D, U$. The vectors do not include speeds earlier than $6$ minutes, because adding such information did not improve prediction quality in our experiments.

\subsubsection{Experiments with Linear Regression} \label{sec:exp_lr}

Linear Regression (LR) assumes the following linear relationship between the response variable $S_0$ and the explanatory variables:
\begin{equation}
    S_0 = \beta_5^S S_5 + \beta_6^S S_6 + \beta_5^U U_5 + \beta_6^U U_6 + \beta_5^D D_5 + \beta_6^D D_6
    \,,
\end{equation}
where $\beta$'s are trainable parameters.
We train this model using Ordinary Least Squares (OLS) to obtain the parameters that minimize the sum of squared differences between observed and predicted values of $S_0$, as detailed in
\cite{weisberg2005applied}.

\subsubsection{Experiments with Gaussian Processes} \label{sec:exp_gp}
Whereas LR is a parametric modeling method, Gaussian Process (GP) regression is non-linear and non-parametric.
GP assumes that the relationship between $S_0$ and the explanatory variables is an unknown function, drawn from a multivariate Gaussian distribution,
\begin{equation}
    \mathcal{N}(\bm{0}, \bm{K})
    \,,
\end{equation}
where for the given $n$ observations, $\bm{K}$ is a $n$-by-$n$ covariance matrix, so that for any two vectors $x_i, x_j$ ($i,j \in 1..n$), $\bm{K}_{i, j}$ expresses the similarity between $x_i$ and $x_j$. 

As a prior distribution, we define the elements of $\bm{K}$ through the commonly used RBF kernel function,
\begin{equation}
    \bm{K}_{i, j} = \exp ( -0.5 \norm{x_i / \tau, x_j / \tau}^2 )
    \,,
\end{equation}
where $\norm{\ }$ denotes the Euclidean distance norm, and $\tau$ is a scaling factor.
To cover a wide range of scaling factors, we experiment with $\tau \in \{0.1, 1, 2, 4, 8, 9, \dots, 16\}$.
Each model is then trained through Maximum Likelihood Estimation (MLE), as detailed in \cite{williams2006gaussian}.

\subsubsection{Experiments with Deep Neural Networks} \label{sec:exp_dnn}

\begin{figure}[t]
    \centering
    \includegraphics[width=0.69\textwidth]{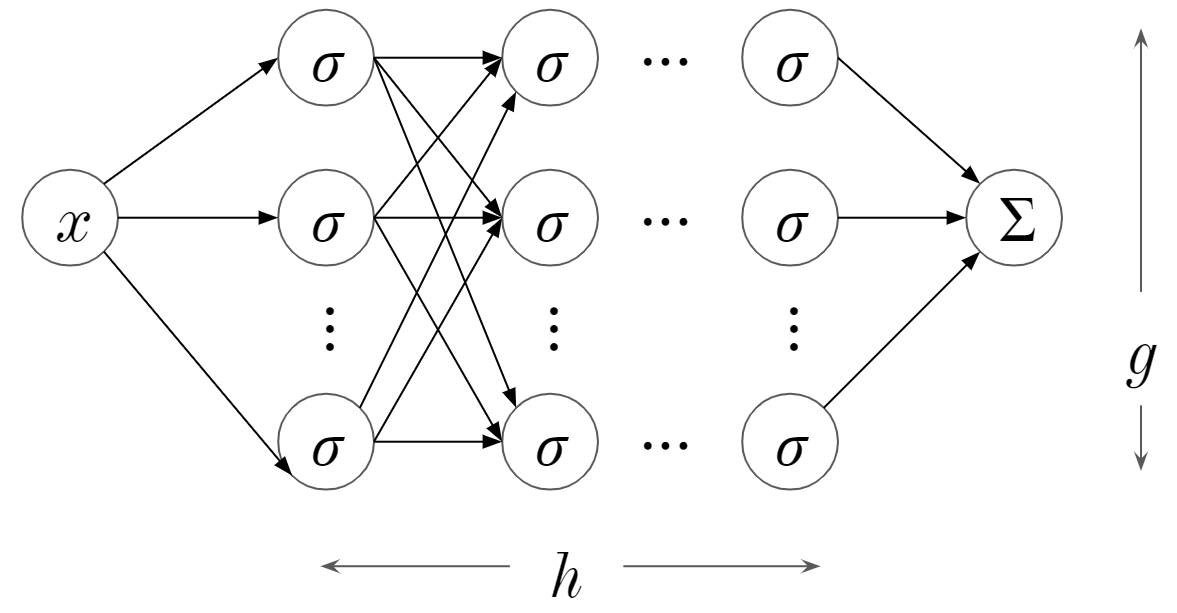}
    \caption{Structure of Deep Neural Networks in our experiments.}
    \label{fig:dnn}
\end{figure}

The Deep Neural Network (DNN) models in our experiments are parametric and non-linear, and are structured as in \fgr \ref{fig:dnn}.
These models pass an input vector $x = \left[ S_5, S_6, U_5, U_6, D_5, D_6 \right]$ through a succession of densely connected hidden layers, and finally output the sum of the results as the predicted value of $S_0$.
Each of the $h$ hidden layers consists of $g$ sigmoid units, each structured as:
\begin{equation}
    \sigma(x) = \frac{1}{1 + \exp(-w^T x)}
    \,,
\end{equation}
where $w$ is a vector of trainable parameters, and $h, g$ are hyper-parameters that we specify below.
For example, if $h=2$ and $g=5$, then the DNN models the relationship between $x$ and $S_0$ as:
\begin{align}
    \nonumber S_0 = 
    &\sigma_{1, 2}\left(
        \left[ \sigma_{1, 1}\left(x\right), \dots, \sigma_{5, 1}\left(x\right) \right]
    \right) 
    + \\
    \nonumber &\sigma_{2, 2}\left(
        \left[ \sigma_{1, 1}\left(x\right), \dots, \sigma_{5, 1}\left(x\right) \right]
    \right)
    + \\
    \nonumber &\cdots +\\
    &\sigma_{5, 2}\left(
        \left[ \sigma_{1, 1}\left(x\right), \dots, \sigma_{5, 1}\left(x\right) \right]
    \right)
    \,,
    \label{eq:dnn}
\end{align}
where $\sigma_{i,j}$ denotes the $i$'th sigmoid in the $j$'th hidden layer.

We have experimented with $h \in \{1, 2\}$ hidden layers, each consisting of $g \in \{5, 10\}$ sigmoids.
Each model is trained with backpropagation for a maximum of $100$ epochs, using Mean Squared Error as loss function, mini batches of size $100$, the Adam optimizer, and a $10\%$ validation split to monitor overfitting.
For more details of this training procedure, we kindly refer the reader to \cite{kingma2014adam}.

\subsubsection{Model Complexity} \label{sec:complexity}

Before proceeding to the experimental results, let us compare the three model types in terms of computational complexity.
Generally speaking, LR is the simplest of the three because it incorporates few parameters. 
DNN and GP, however, are not directly comparable: GP is non-parametric and scales cubically with the number of input vectors, whereas the complexity of fitting DNN models depends on the number of epochs and mini-batches.
The complexity of prediction, however, is quite similar across all model types, i.e., it takes roughly the same time to generate predictions from any of the models once trained. 

\section{Results} \label{sec:results}

We now provide the results of the experiments that we constructed in Section \ref{sec:case_study}. 
The results will show that for a wide range of incident cases, the information in IVMS signals suffices for $M_{abnormal}$ to reduce prediction errors significantly and timely. 
While analyzing the results, we also provide some insights into the behavior of traffic under incidents, and examine how prediction quality deteriorates if model adaptation is not performed.

\subsection{Real-Time Performance of QTIP} \label{sec:realtime}

We executed all experiments on a server with $16$ GB of memory and an Intel i7-2600 CPU, clocked at 3.40 GHz. 
It took QTIP at most $45\ s$ to simulate $15\ min$, for any single what-if simulation. 
Once QTIP has executed all what-if simulations, it required only a few more seconds to fit an LR model $M_{abnormal}$ on their results. 
Hence when all what-if simulations are executed in parallel, QTIP can yield an adapted model $M_{abnormal}$ within $1\ min$, just in time for handling the first critical minutes of the incident. 
It should also be noted that recently developed methods for GPU acceleration of micro-simulations \citep{heywood2018gpu} may further cut down these running times.

\subsection{Comparison of Predictive Performance} \label{sec:compare_performance}

\begin{table}[t]
    \caption{Best performing model of each type. Values closer to zero are better, best values are highlighted in bold.}
    \label{tbl:qtip_instances}
    \centering
    \begin{tabular}{l|c|c}
    \Xhline{2\arrayrulewidth}
    \bfseries  & \multicolumn{2}{c}{\bfseries Mean RMSE} \\
    \bfseries Model Type & \bfseries Lanes Known &  \bfseries Lanes Unknown \\
    \hline
    GP $\tau=8$    &  13.928        &  23.676 \\
    DNN $h=2, g=10$  &  6.328         &  9.145 \\
    LR               &  \bm{$6.290$}  &  \bm{$8.878$} \\
    \Xhline{2\arrayrulewidth}
    \end{tabular}
\end{table}

For each model type among GP, DNN, and LR, Table \ref{tbl:qtip_instances} summarizes the predictive quality of the best performing model as measured through mean RMSE (Eq. \ref{eq:rmse}) over all experiments. 
As may be expected, all models perform better when blocked lanes are known than otherwise. 

LR is the best performing model type when lanes are either known or unknown.
GP performs much worse than both LR and DNN, possibly because of inappropriate choice of kernel function or overfitting to train data.
The poor fit of GP may thus be alleviated through, e.g., incorporating a periodic kernel function or applying regularization techniques \citep{roberts2013gaussian}. 
The predictive performance of DNN is a little worse than that of LR, as we next explain.

The small differences in predictive performance between LR and DNN could be caused by a number of reasons.
First, the dataset at hand may conform better to linear (LR) than non-linear (DNN) patterns, although this could change if more explanatory variables are added or if substituting for a larger dataset.
Second, while the training methods of both models are based on stochastic optimization, training a DNN through backpropagation typically requires more delicate tuning than training LR through OLS.
It could thus be that finer selection of DNN hyper-parameters, such as via Bayesian Optimization \citep{snoek2012practical}, would yield a better performing DNN model.
Finally, the DNN models can be made more powerful through a change of structure, e.g., by stacking Recurrent units for time series data \citep{ho2002comparative}, thus adding memory capabilities that LR does not possess.

While further improvement of the models is intriguing in itself, it is nevertheless superfluous for the objectives of this paper.
First, LR, DNN, and GP are commonly used model types for traffic modeling, and they demonstrate (Table {\ref{tbl:qtip_instances}}) that higher precision of incident location indeed yields better predictions across all model types.
Second and more importantly, the particular choice of model types is less important for proof-of-concept purposes, as the results we next present successfully convey the fundamental message of this paper: that model degradation under incidents can be mitigated through just-in-time, simulation-based adaptation.
Third, as explained in Section {\ref{sec:motivation}}, the main source of complexity for modeling under incidents is, in any case, the sudden change of variable correlation structure, regardless of model form.
We thus defer model improvements to future work (Section {\ref{sec:future}}), and focus next on the predictive performance of QTIP with LR, the best performing model.

\subsection{Closer Look at Several Representative Incident Scenarios}

\floatsetup[figure]{style=plain, subcapbesideposition=center}

\begin{figure*}[t]
    \centering
    \sidesubfloat[][]{
        \includegraphics[trim={4cm, 3cm, 4cm, 1cm}, clip, width=0.45\textwidth]{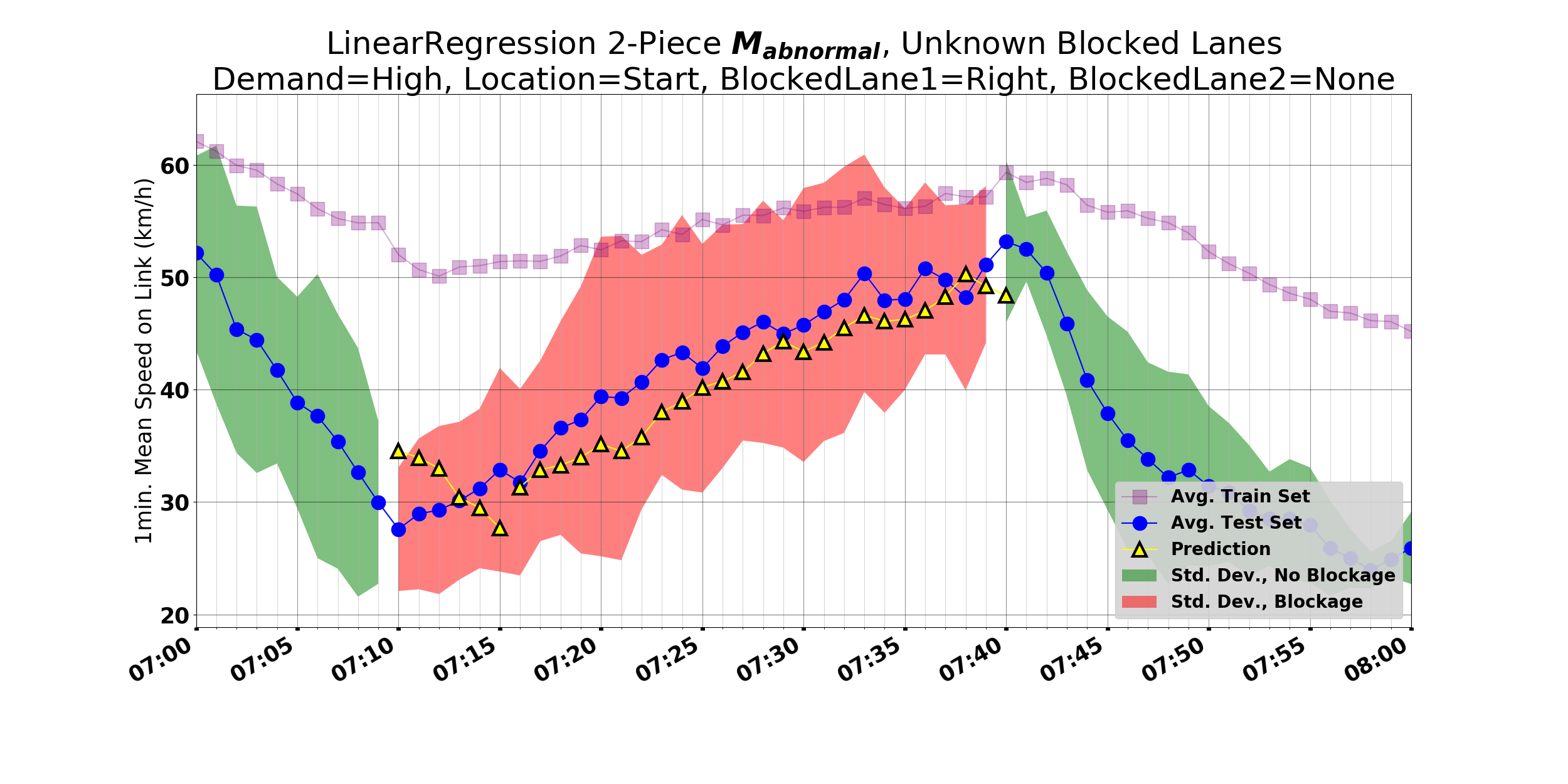}
        \label{fig:unknown1}
    }
    \hfil
    \sidesubfloat[][]{
        \includegraphics[trim={4cm, 3cm, 4cm, 1cm}, clip, width=0.45\textwidth]{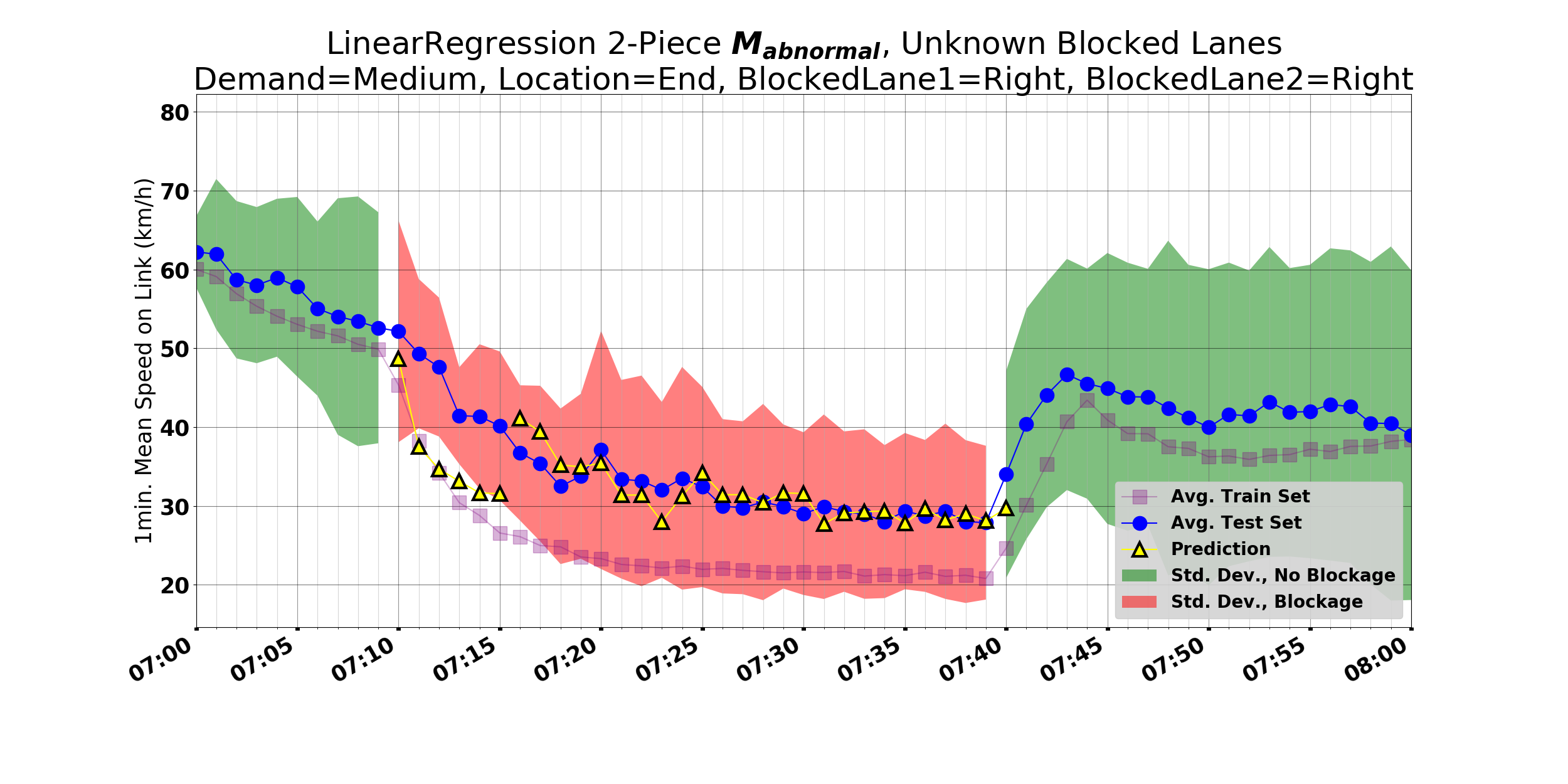}
        \label{fig:unknown2}
    }
    \hfil
    \sidesubfloat[][]{
        \includegraphics[trim={4cm, 3cm, 4cm, 1cm}, clip, width=0.45\textwidth]{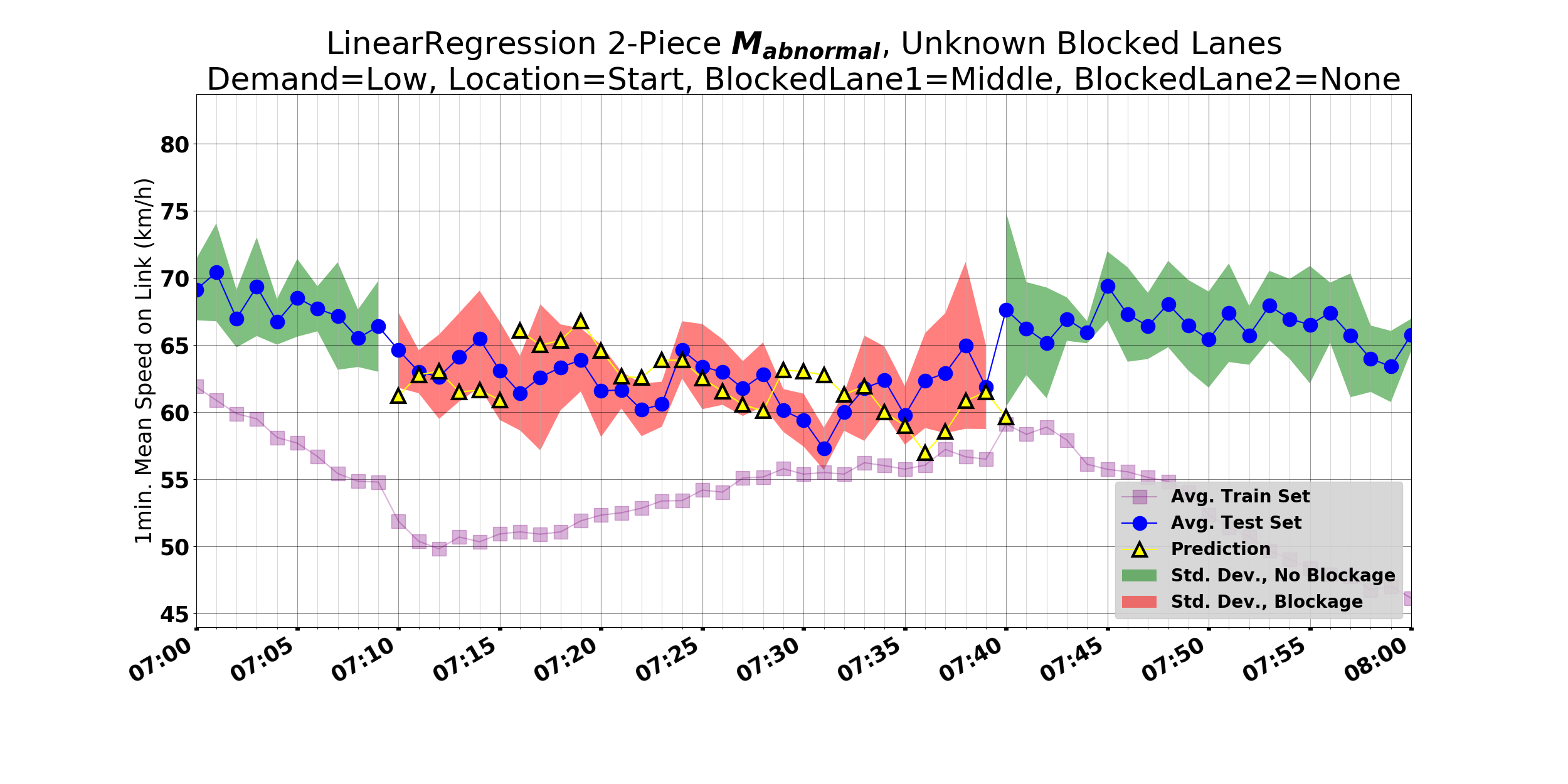}
        \label{fig:unknown3}
    }
    \hfil
    \sidesubfloat[][]{
        \includegraphics[trim={4cm, 3cm, 4cm, 1cm}, clip, width=0.45\textwidth]{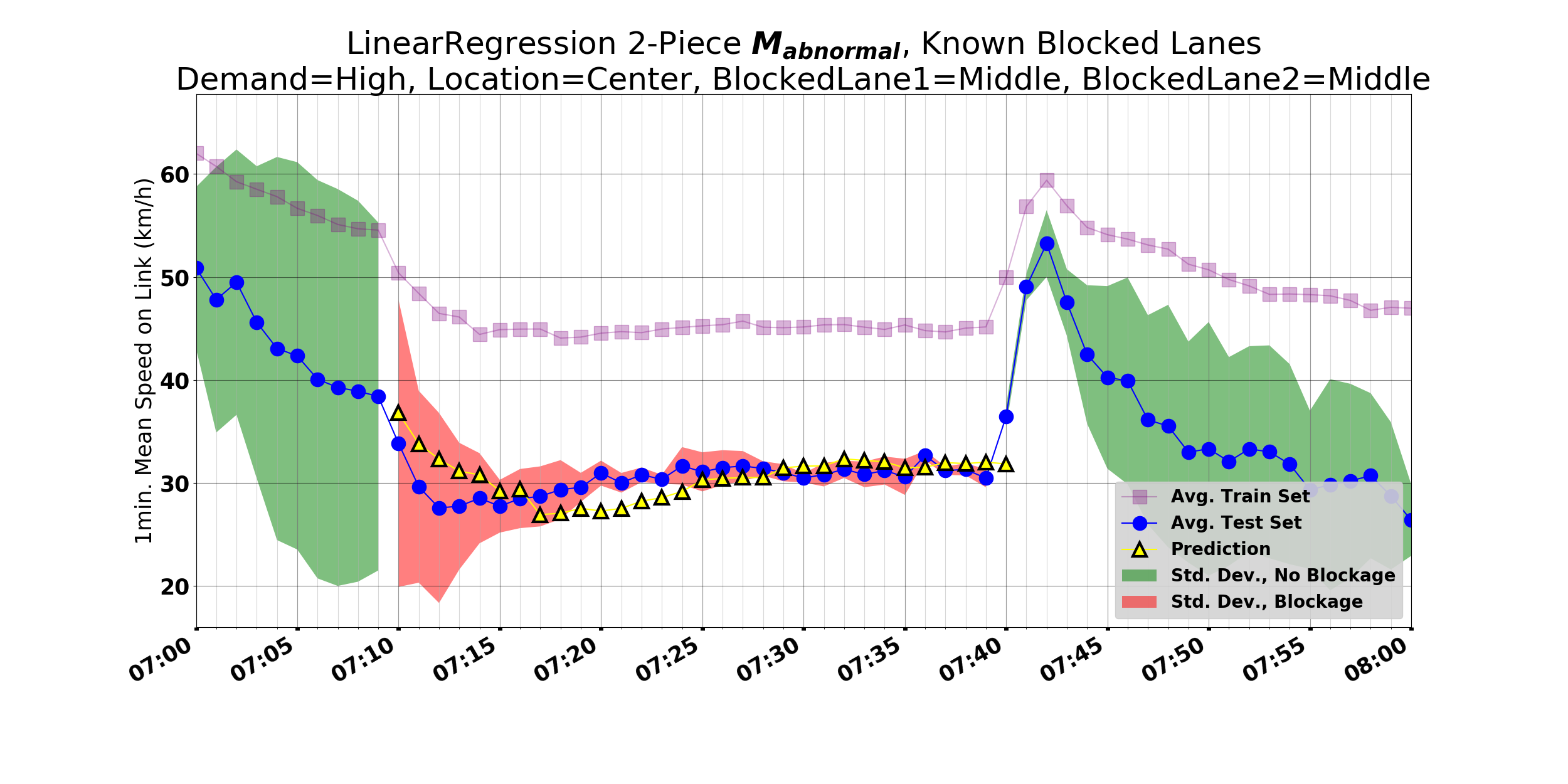}
        \label{fig:known1}
    }
    \hfil
    \sidesubfloat[][]{
        \includegraphics[trim={4cm, 3cm, 4cm, 1cm}, clip, width=0.45\textwidth]{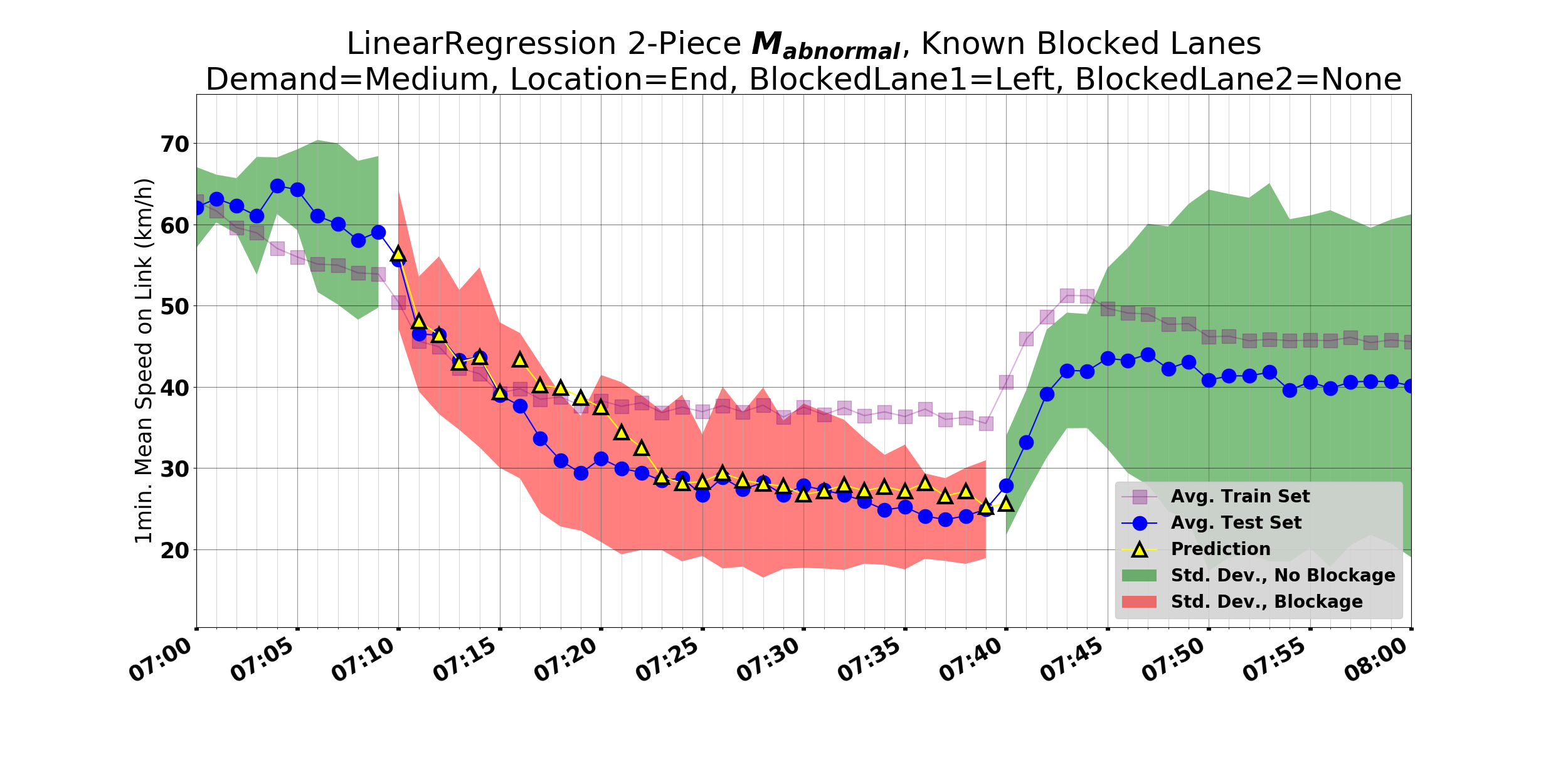}
        \label{fig:known2}
    }
    \hfil
    \sidesubfloat[][]{
        \includegraphics[trim={4cm, 3cm, 4cm, 1cm}, clip, width=0.45\textwidth]{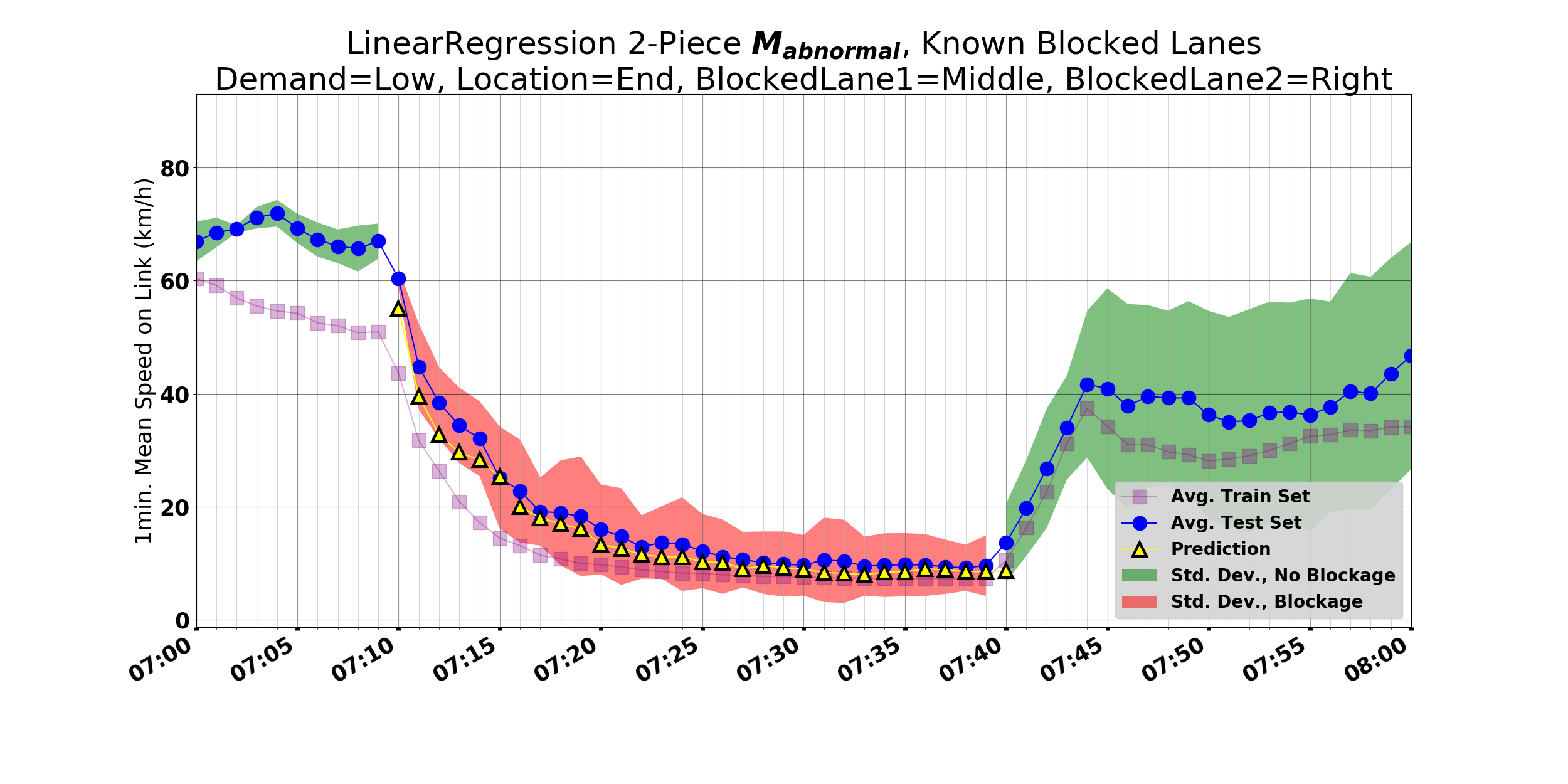}
        \label{fig:known3}
    }    
    \caption{Representative examples of behavior of mean speed in Study link $S$ under different incident scenarios, and performance of $M_{abnormal}$.}
    \label{fig:example_scenarios}
\end{figure*}

While analyzing the results of our experiments, we have observed several different behaviors of the mean speed under incidents. We now illustrate these behaviors by selecting and visualizing several representative scenarios, for both cases of known and unknown blocked lanes. For each representative scenario, we plot in \fgr \ref{fig:example_scenarios} the time series of $1\ min$ ground-truth speed, averaged over all ground-truth simulations of that scenario. For further analysis, we also plot the predictions of $M_{abnormal}$ when applied to each of these averaged time series.

First, we see in \fgr \ref{fig:example_scenarios} that mean speed in the Study Link $S$ mostly drops after the onset of an incident (\ref{fig:unknown2}, \ref{fig:known1}, \ref{fig:known2}, \ref{fig:known3}),
but less frequently, it can also gradually increase (\ref{fig:unknown1}). This exceptional phenomenon occurs when the road blocks appear at the start of the link, so that vehicles which eventually enter the link can then flow freely. We also see that the standard deviation of mean speed typically exhibits more variation when only $1$ lane is blocked (\ref{fig:unknown1}, 
\ref{fig:unknown2},
\ref{fig:known1},
\ref{fig:known2}) than when $2$ different lanes are blocked (\ref{fig:known3}). This is because $2$ blocked lanes bring about a level of congestion, such that queued vehicles travel in similarly low speeds, whereas speeds are more variable when only $1$ lane is blocked. Finally, we see that the predictions of $M_{abnormal}$ often come close to the actual mean speed during the incidents. Indeed, we next show that $M_{abnormal}$ mostly outperforms $M_{ordinary}$ in incident scenarios.

\subsection{Model for Ordinary Conditions Degrades under Incidents} \label{sec:degradation}

Let us now measure what happens if QTIP is unavailable, so that only $M_{ordinary}$ is used for traffic prediction. To this end, we compare the performance of LR $M_{ordinary}$ on incident-free simulations vs. incident simulations. The performance measurements we use are: Mean Signed Deviation (MSD), Mean Absolute Error (MAE), and Rooted Mean Squared Error (RMSE). 
These are defined as following for any set of vectors 
$V = \{v_1, \dots, v_N\}$
, where for 
$v_i \in V$, $r_i$ 
is the model estimation minus the actual value of the response value.
\begin{align}
\text{MSD} &= 
\left( r_1 + \cdots r_N \right) / N \,,\\
\text{MAE} &= 
\left( |r_1| + \cdots |r_N| \right) / N \,,\\
\label{eq:rmse} \text{RMSE} &= 
\left( \left( r_1^2 + \cdots r_N^2 \right) / N \right)^{0.5} \,.
\end{align}

\begin{figure*}[t]
    \centering
    \subfloat[]{
        \includegraphics[trim={2cm, 2cm 2cm, 2cm}, clip, width=0.31\textwidth]{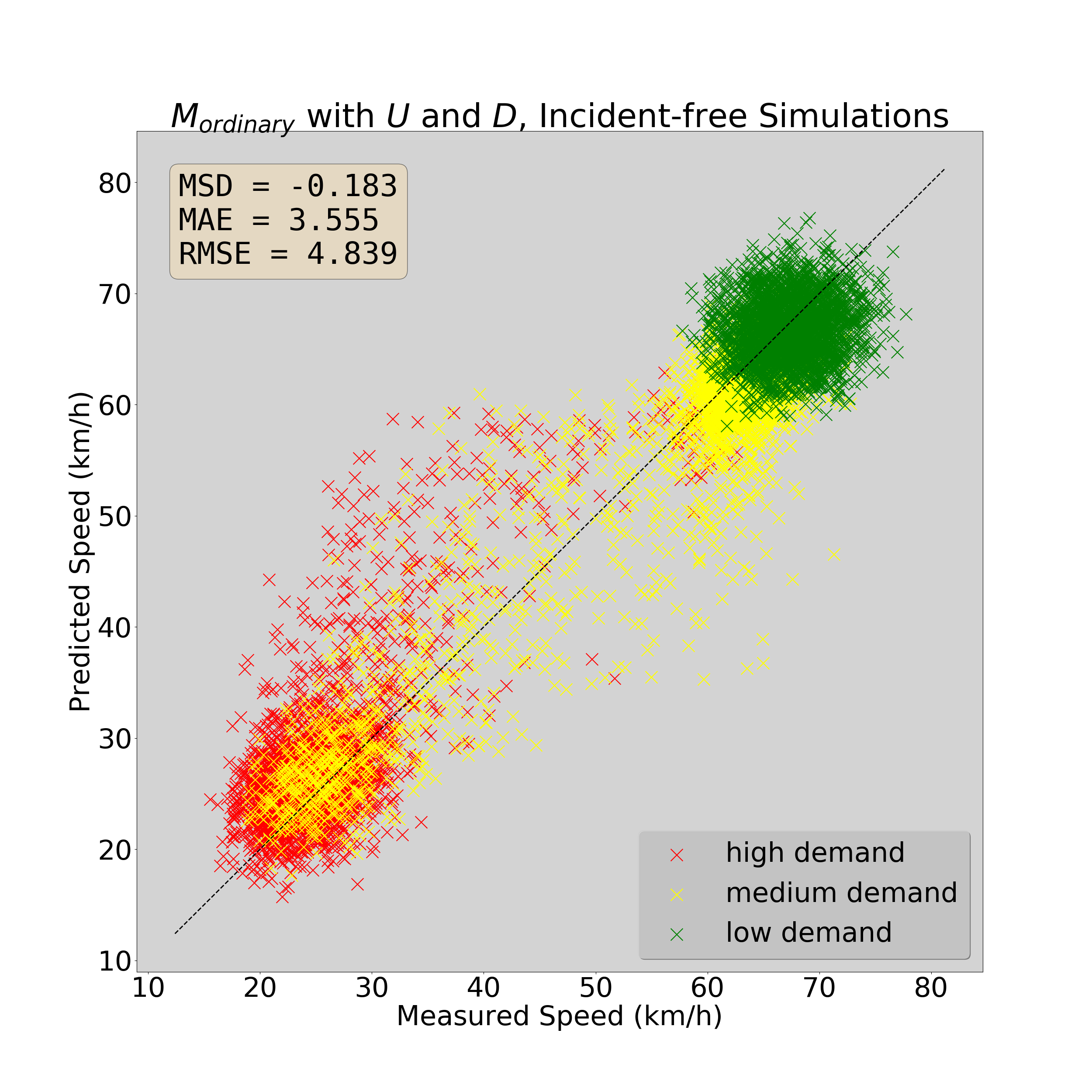}
        \label{fig:m_ordinary_with_u_d}
    }
    \hfil
    \subfloat[]{
        \includegraphics[trim={2cm, 2cm 2cm, 2cm}, clip, width=0.31\textwidth]{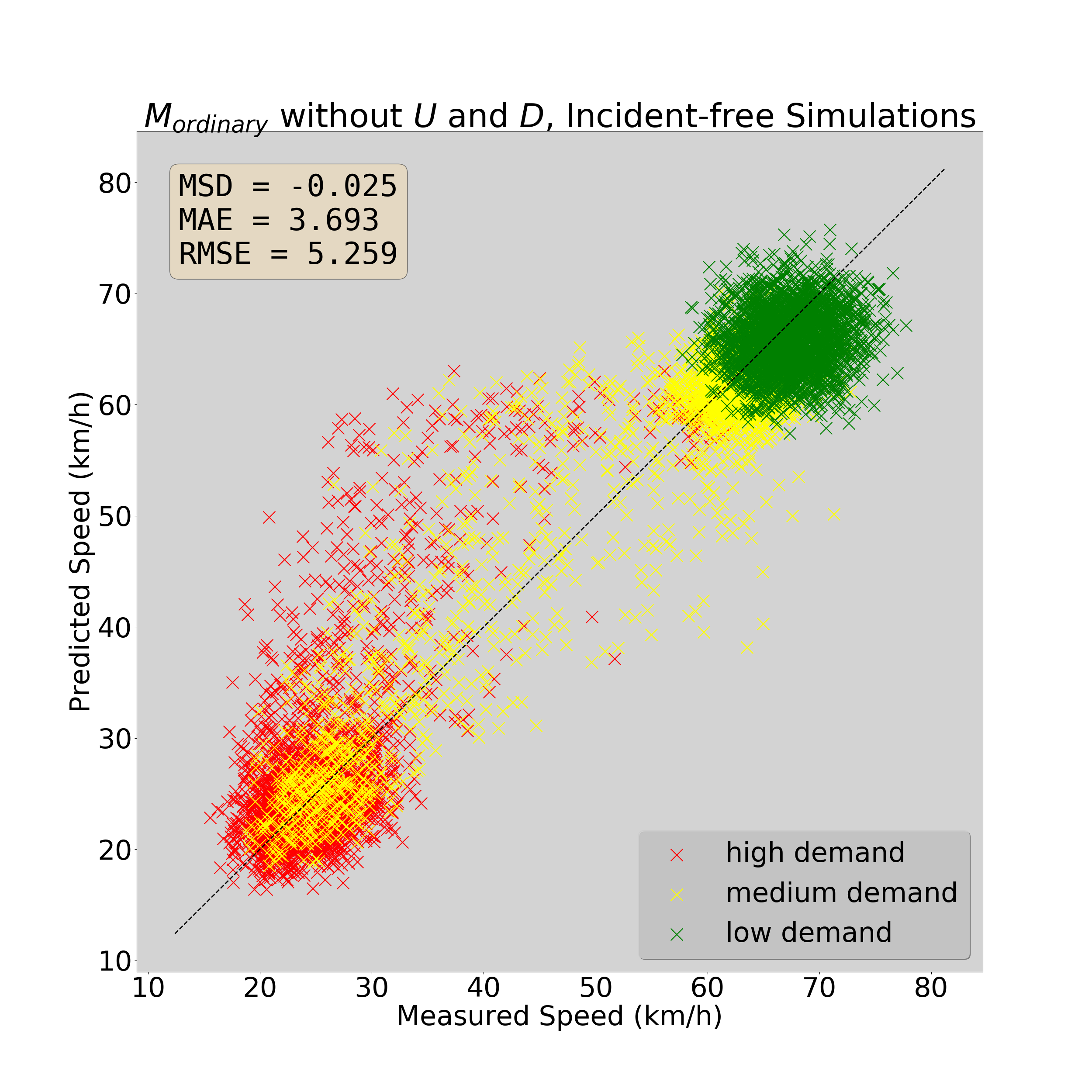}
        \label{fig:m_ordinary_without_u_d}
    }
    \subfloat[]{
        \includegraphics[width=0.315\textwidth, trim={4cm, 5cm, 3cm, 5cm}, clip,]{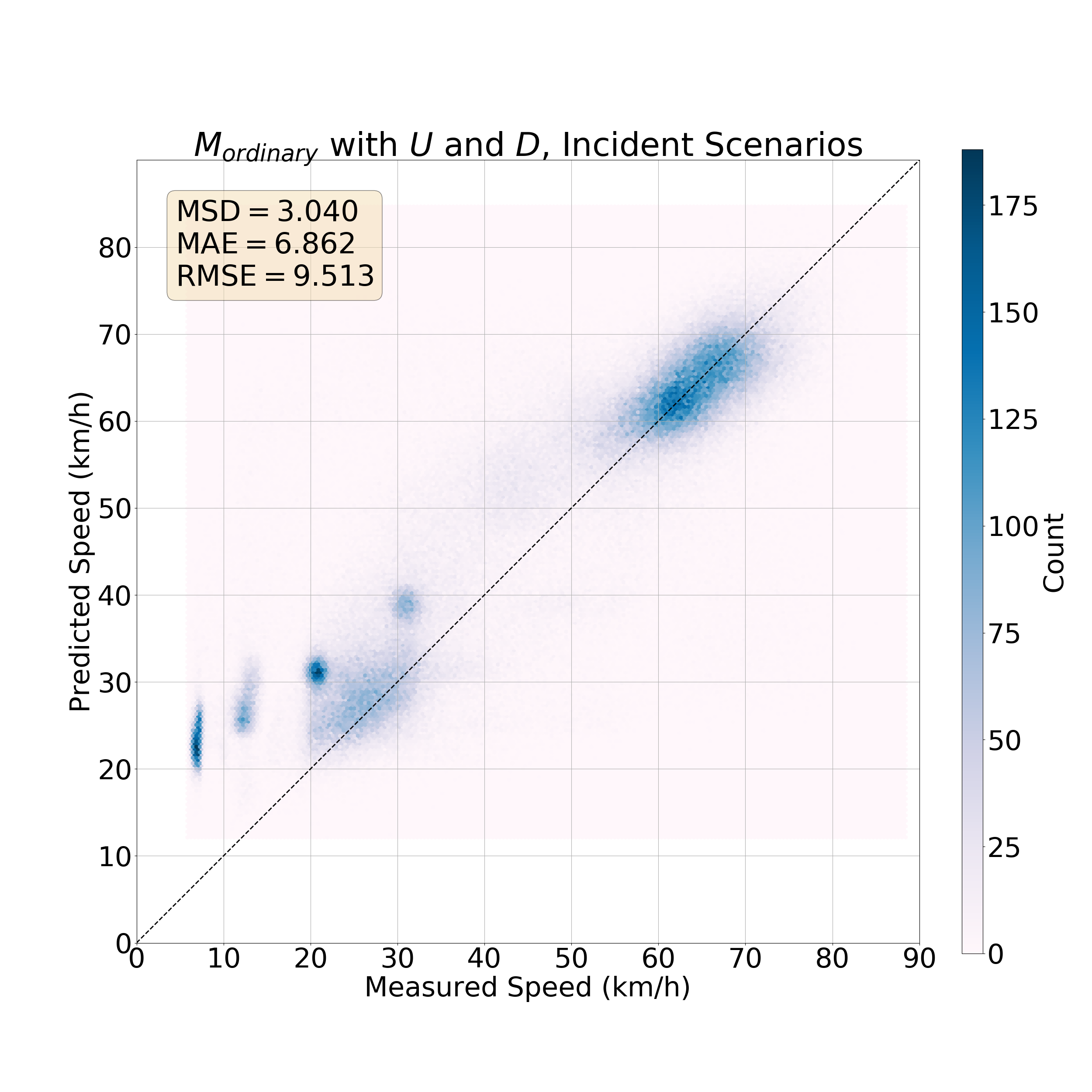}
        \label{fig:degradation_scatter}
    }
    \caption{Performance of $M_{ordinary}$ under normal and abnormal traffic conditions in our experiments. All plots show predicted vs. ground-truth mean speed, such that accurate predictions lie on the diagonal line. Under normal conditions, $M_{ordinary}$ performs better when its input contains information about the uplink $U$ and downlink $D$ (\ref{fig:m_ordinary_with_u_d}) than without this information (\ref{fig:m_ordinary_without_u_d}). For incident conditions, \ref{fig:degradation_scatter} shows the density of scattered points. The predictive performance of $M_{ordinary}$ degrades considerably under incidents, and it tends to over-estimate the mean speed, as also reflected in the Mean Signed Deviation (MSD).}
    \label{fig:normal_lr}
\end{figure*}

For incident-free simulations, we apply $10$-fold cross validation, and obtain \fgr \ref{fig:normal_lr}, which illustrates the prediction accuracy of $M_{ordinary}$ for each vector from incident-free simulations. We see that $M_{ordinary}$ is mostly accurate in predicting speeds under incident-free conditions, as we might expect, because traffic in highways flows in a rather regular manner when no disruptions occur.

Next, we measure how worse $M_{ordinary}$ performs on incident conditions, for which it was not trained. The results are summarized in Table \ref{tbl:m_ordinary_performance}, and we see that $M_{ordinary}$ deteriorates in comparison with its performance under normal conditions. The deterioration occurs whether or not $M_{ordinary}$ has access to information about uplink $U$ and downlink $D$. For the former option, \fgr \ref{fig:degradation_scatter} depicts the tendency of $M_{ordinary}$ to overestimate the actual mean speed in study link $S$.

\begin{table}[t]
    \caption{Performance of $M_{ordinary}$ under normal vs. incident conditions}
    \label{tbl:m_ordinary_performance}
    \centering
    \begin{tabular}{c|c|c|c|c}
    \Xhline{2\arrayrulewidth}
    \bfseries Conditions & \bfseries $U$ and $D$ & \bfseries MSD & \bfseries MAE & \bfseries RMSE \\
    \hline 
    \multirow{ 2}{*}{Normal} & with & $-0.183$ & $3.555$ & $4.839$ \\
    & w/o & $-0.025$ & $3.693$ & $5.259$ \\
    \hline
    \multirow{ 2}{*}{Incident} & with & $3.040$ & $6.862$ & $9.513$ \\
    & w/o & $0.440$ & $5.913$ & $9.085$ \\
    \Xhline{2\arrayrulewidth}
    \end{tabular}
\end{table}

\begin{figure}[t]
    \centering
    \includegraphics[width=\textwidth, trim={4cm, 3cm, 4cm, 1cm}, clip]{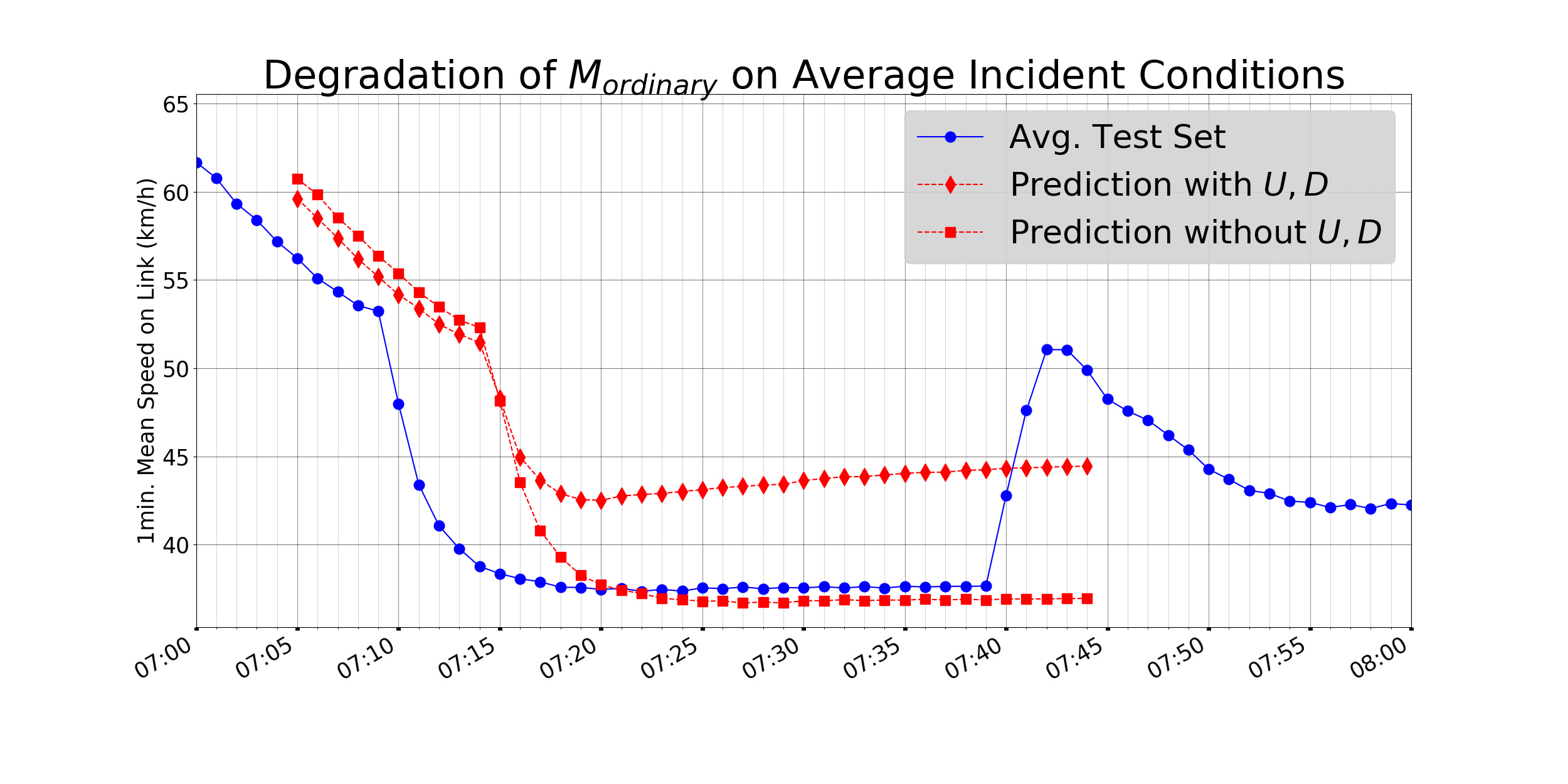}
    \caption{Degradation of $M_{ordinary}$ when applied to the average incident time series, either with or without input from uplink $U$ and downlink $D$.}
    \label{fig:degradation_ts}
\end{figure}

To further reason about this deterioration, we next visualize how $M_{ordinary}$ performs on an the average incident time series, where each lag is the $1\ min$ mean speed, averaged over all ground-truth simulations. \fgr \ref{fig:degradation_ts} shows how $M_{ordinary}$ fails to predict the average incident time series, either with or without information about $U$ and $D$. In the former case, the predictions of $M_{ordinary}$ on the average incident time series do not even converge to the typical speed at the second phase of the incidents. Upon examining simulations output, it seems that this lack of convergence is caused by an increase in mean speed in $D$ while congestion forms in $S$.

In fact, we also see in \fgr \ref{fig:degradation_ts} the three typical phases of the mean speed under road incidents, as following. The first phase immediately follows the onset of the incident, at which time the mean speed drops sharply for a few minutes. In the second phase, the mean speed stabilizes, while the incident is still in place. The third phase immediately follows the clearing of the incident, at which time the mean speed increases sharply for a few minutes, resuming the trend it had just before the incident.

\subsection{Improvement of Predictive Performance with QTIP} \label{sec:improvement}

\begin{figure*}[t]
    \centering
    \includegraphics[width=\textwidth,trim={3px 3px 3px 3px},clip]{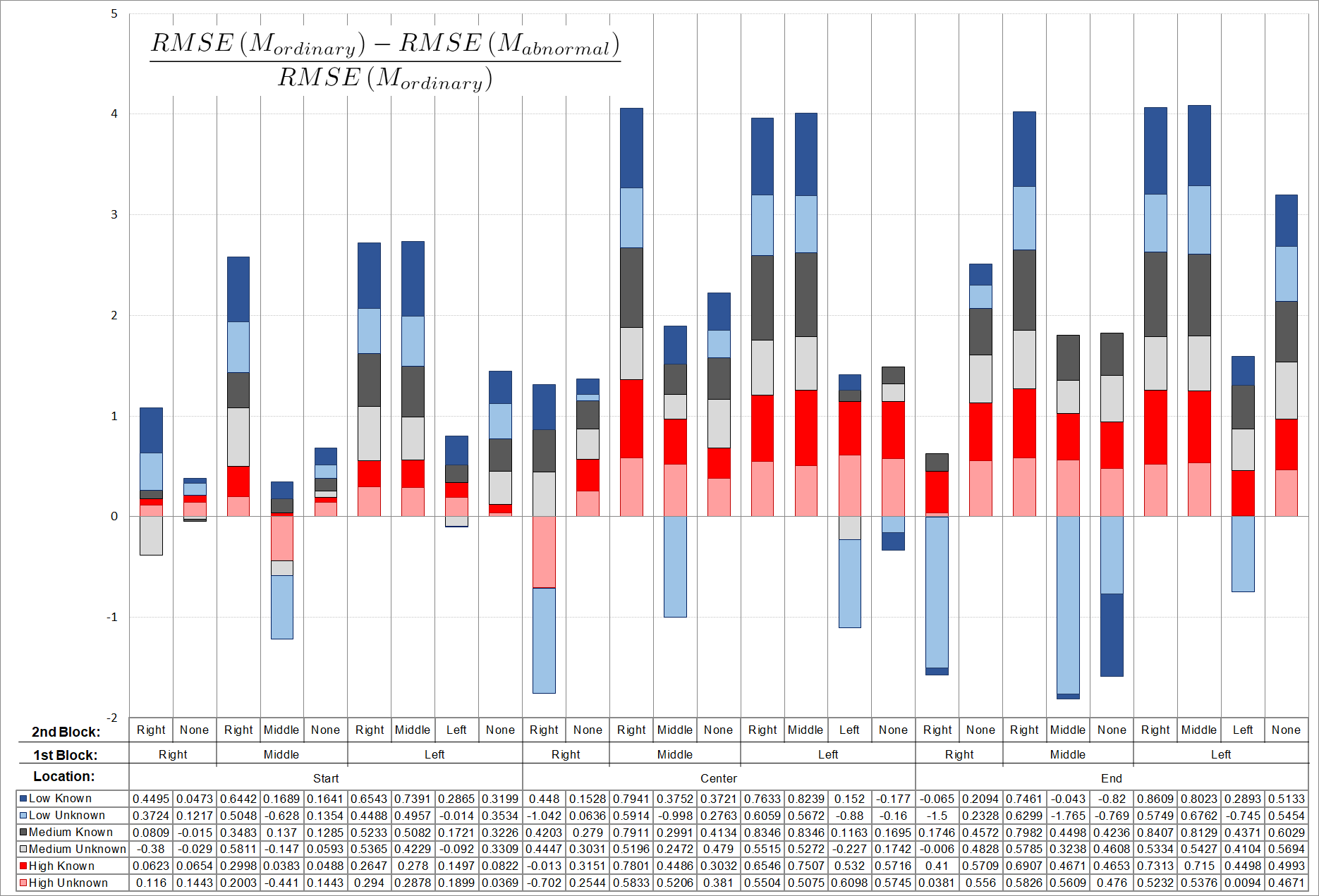} 
    \caption{Relative improvement of RMSE of $M_{abnormal}$ over $M_{ordinary}$ in the first $6$ minutes of incidents, when QTIP either knows or does not know which lanes are blocked. All models use Linear Regression, and higher values correspond to greater improvement.}
   \label{fig:rmse_improvement}
\end{figure*}

We have shown that $M_{ordinary}$ deteriorates significantly under incident conditions, which further expresses the need for just-in-time model adaptation. Therefore, we now measure the performance gain when using the adapted $M_{abnormal}$ instead of $M_{ordinary}$ in the first $6$ minutes of each incident scenario. To this end, we calculate the relative RMSE improvement for each scenario, as:
\begin{align}
    \frac
    {\text{RMSE}\left(M_{ordinary}\right) - \text{RMSE}\left(M_{abnormal}\right)}
    {\text{RMSE}\left(M_{ordinary}\right)}
    \,,
\end{align}
where $\text{RMSE}(M)$ is the RMSE of predictions of model $M$ for the corresponding scenario. Positive values correspond to lower prediction error of $M_{abnormal}$ vs. $M_{ordinary}$, namely better performance of $M_{abnormal}$.

\fgr \ref{fig:rmse_improvement} visualizes the relative RMSE improvement for each incident scenario. We see that $M_{abnormal}$ mostly outperforms $M_{ordinary}$, whether or not QTIP knows the exactly blocked lanes. Moreover, we see that when QTIP knows which lanes are blocked, the prediction quality of $M_{abnormal}$ considerably increases. Averaging all the values in \fgr \ref{fig:rmse_improvement}, we obtain that the mean relative RMSE improvement over all incident scenarios is $28.74$\%.

We also see in \fgr \ref{fig:rmse_improvement} a few exceptional cases, in which $M_{ordinary}$ outperforms $M_{abnormal}$. Most of these cases have the following in common: two adjacent vehicles block the same lane, demand is low, and location precision is low. Such circumstances are illustrated in \fgr \ref{fig:unknown3}, where we see that traffic disruption is then rather minor, so that the behavior of the mean speed remains rather stable when the incident occurs. Hence in such circumstances, on one hand, the uninformed $M_{ordinary}$ performs well. On the other hand, QTIP does not know that the two distress signals originate from vehicles on the same lane, and so generates what-if simulations also for two different blocked lanes. Consequently, in such circumstances, QTIP trains $M_{abnormal}$ to predict a disruption greater than actual.

\subsection{Transfer Learning} \label{sec:transfer-learning}

In the above experiments, $M_{abnormal}$ is fit afresh for each scenario.
Nevertheless, as $M_{ordinary}$ incorporates knowledge about historical traffic, it could be beneficial to give $M_{abnormal}$ access to this knowledge during training.
In other words, $M_{abnormal}$ could possibly benefit from some form of transfer learning \citep{pan2009survey}.

To explore this possibility, we next carry out additional experiments, where the linear coefficients of $M_{abnormal}$ are based on the coefficients of $M_{ordinary}$.
In these additional experiments, we again use the same $81$ incident scenarios as in Section \ref{sec:common}, when QTIP either knows or does not know the precisely blocked lanes.
This time, however, we use \emph{Bayesian Inference} \citep{pereira2019mobility} to obtain $\vecbeta \in \mathbb{R}^k$, the linear coefficients of $M_{abnormal}$.

\subsubsection{Bayesian Setup}

In each experiment, the \emph{prior} on the coefficients of $M_{abnormal}$ is a multivariate Gaussian,
\begin{align}
    p(\vecbeta) &= \gaussian{\vecbeta}{\vecmu}{\sigma_\beta^2 \mat{I}_k}
    \,,
\end{align}
where $\vecmu \in \mathbb{R}^k$ is the coefficients of $M_{ordinary}$, $\mat{I_k}$ is the $k \times k$ identity matrix, and $\sigma_\beta$ is a hyper-parameter.
Hence without further evidence about incidents, $M_{abnormal}$ is initially similar to $M_{ordinary}$.
When such evidence is given as features $\mat{X} \in \mathbb{R}^{n \times k}$ and corresponding observations $\vec{y} \in \mathbb{R}^n$, the \emph{likelihood} of the observations is
\begin{align}
    p(\vec{y} \mid \mat{X}, \vecbeta, \sigma_y) 
    &=
    \gaussian{\vec{y}}{\mat{X}\vecbeta}{\sigma_y^2 \mat{I}_n}
    \,,
\end{align}
where $\sigma_y$ is another hyper-parameter.
In this Section, we fix $\sigma_\beta = 1$ and $\sigma_y = 1$; we have also experimented with significantly higher and lower values of $\sigma_\beta$ and $\sigma_y$, but obtained no noticeable change in results.

By Eqs. (2.113)--(2.117) in \citet[Section~2.3]{bishop2006pattern}, the \emph{posterior} on $M_{abnormal}$ coefficients is
\begin{align}
    p (\vecbeta \mid \vec{y}) 
    &= 
    \biggaussian{\vecbeta}
    {\mat{\Sigma}
        \Big\{ 
            \mat{X}^{\mbox{\scriptsize T}} 
            \left( \sigma_y^{-2} \mat{I}_n \right)
            \vec{y} 
            + 
            \left(\sigma_\beta^{-2} \mat{I}_k \right)
            \vecmu 
        \Big\}
    }
    {\mat{\Sigma}}
    \,,
    \label{eq:posteriorbeta}
\end{align}
where $\mat{\Sigma} \in \mathbb{R}^{k \times k}$ is
\begin{align}
    \mat{\Sigma} &= 
    \left( 
    \sigma_\beta^{-2} 
    \mat{I}_k 
    + 
    \mat{X}^{\mbox{\scriptsize T}}
    \left(\sigma_y^{-2} \mat{I}_n\right)
    \mat{X} 
    \right)^{-1}
    \,.
    \label{eq:Sigma}
\end{align}
We thus use the posterior mean of Eq. \ref{eq:posteriorbeta} as the fitted coefficients of $M_{abnormal}$.

\subsubsection{Experiments and Results}

The experiments in this Section allow us to study how an increasing amount of simulated incident information affects Bayesian $M_{abnormal}$ vs. freshly fit $M_{abnormal}$.
For each incident scenario $i$, we first independently generate $5$ ground truth simulations of the incident, which we later use for evaluating the models.
Then for each $i$ and $j=1, 2, \dots, 10$, we have QTIP independently generate $j$ what-if simulations, on which we train each model.
Because each training simulation involves stochastic variations, we repeat the experiment $30$ times for each $i, j$, evaluate the RMSE of each model's predictions vs. the ground truth observations, and average the $30$ evaluations.

\begin{figure*}
    \centering
    \includegraphics[width=\textwidth]{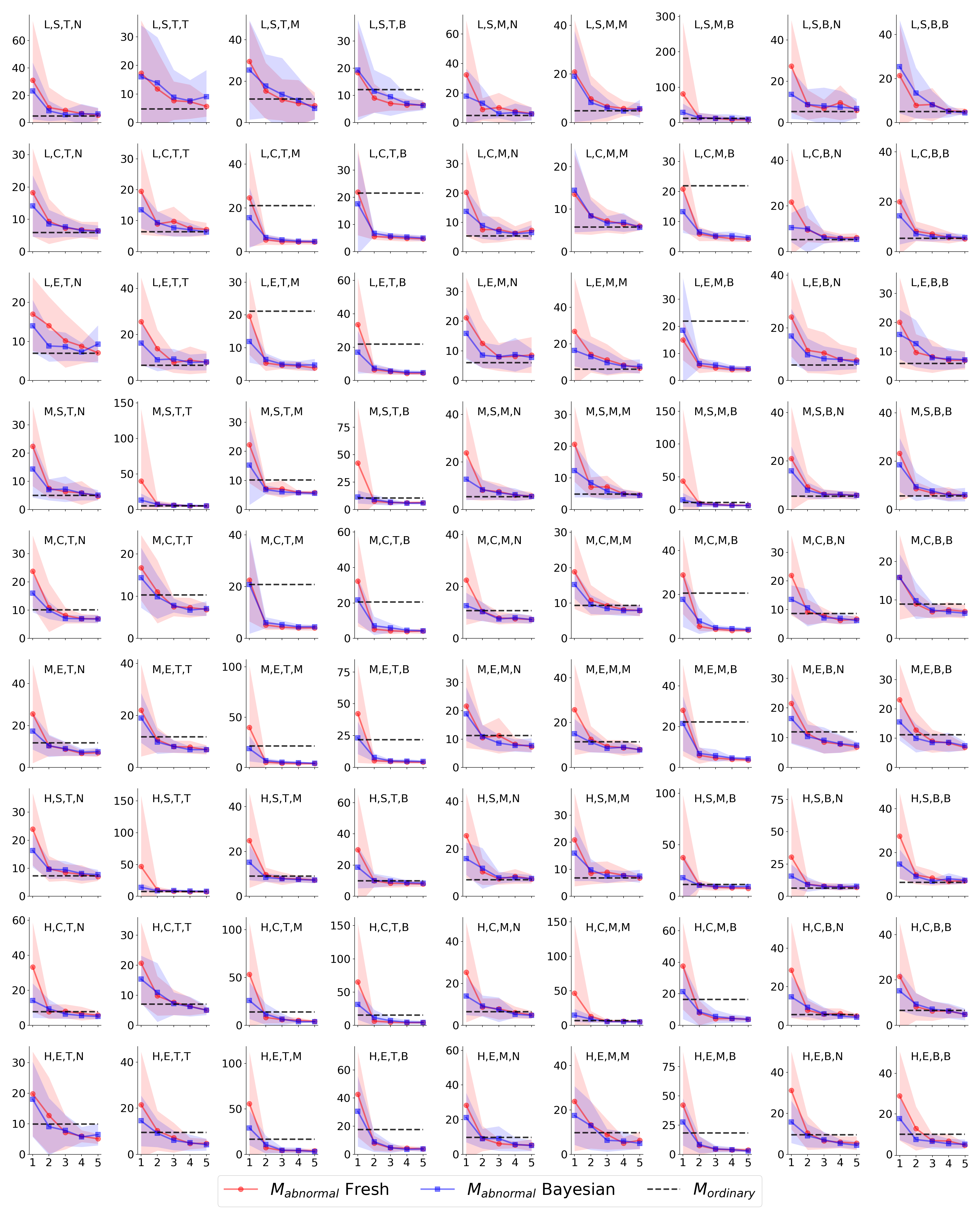}
    \caption{
    Average RMSE (km/h, vertical axis) as the number of training simulations increases (horizontal axis), when QTIP \emph{knows} which lanes are blocked.
    The shaded areas are $\pm 1$ standard deviation around the average.
    Titles pertain to incident scenarios, as: demand level (\textbf{H}igh, \textbf{M}edium, \textbf{L}ow), location on link (\textbf{S}tart, \textbf{C}enter, \textbf{E}nd), 1st and 2nd blocked lanes (\textbf{T}op, \textbf{M}iddle, \textbf{B}ottom, \textbf{N}one).
    }
    \label{fig:bayesian-rmse-known}
\end{figure*}

\begin{figure*}
    \centering
    \includegraphics[width=\textwidth]{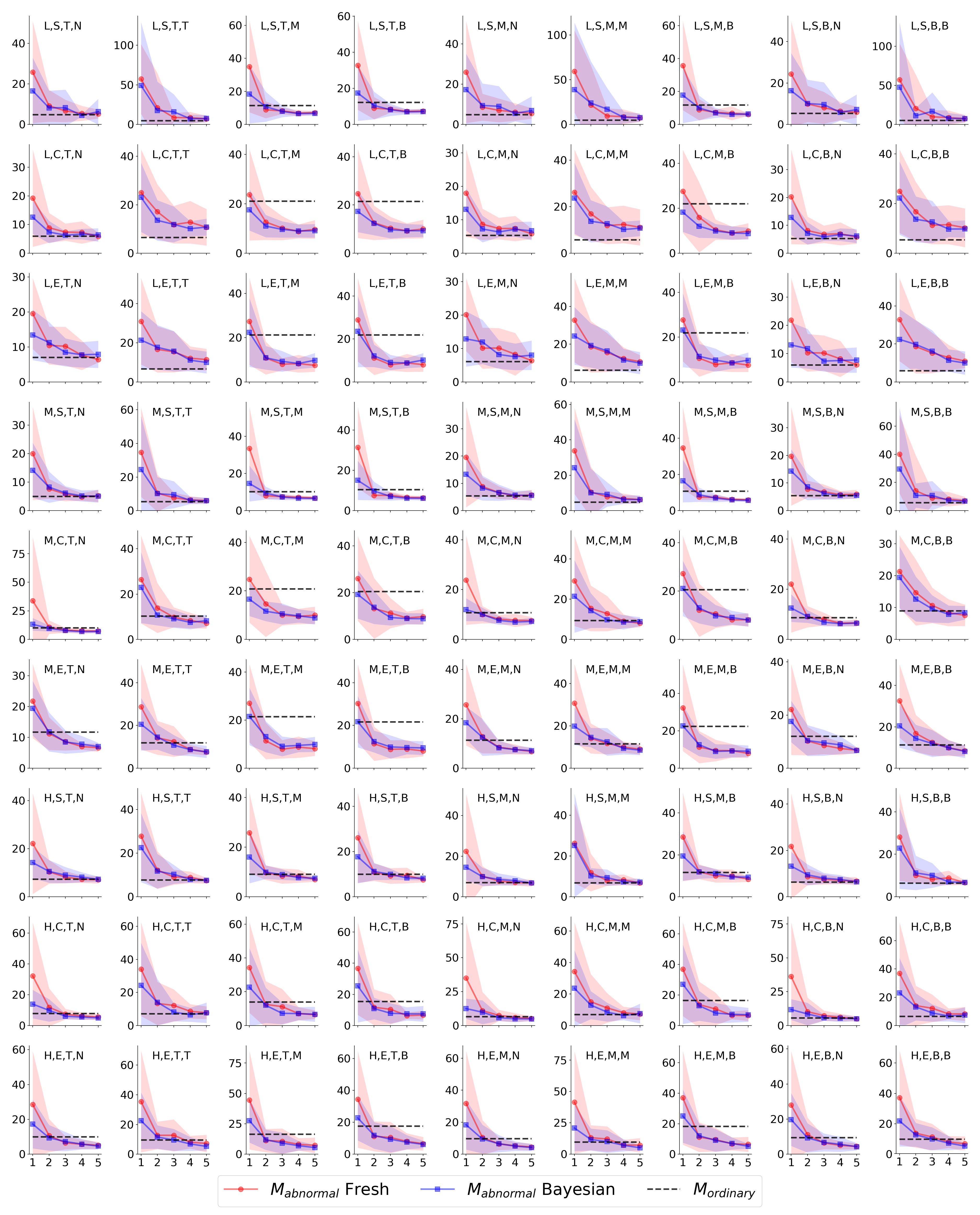}
    \caption{
    Average RMSE (km/h, vertical axis) as the number of training simulations increases (horizontal axis), when QTIP \emph{does not know} which lanes are blocked.
    The shaded areas are $\pm 1$ standard deviation around the average.
    Titles pertain to incident scenarios, as: demand level (\textbf{H}igh, \textbf{M}edium, \textbf{L}ow), location on link (\textbf{S}tart, \textbf{C}enter, \textbf{E}nd), 1st and 2nd blocked lanes (\textbf{T}op, \textbf{M}iddle, \textbf{B}ottom, \textbf{N}one).
    }
    \label{fig:bayesian-rmse-unknown}
\end{figure*}

\fullfgrref{fig:bayesian-rmse-known} depicts the results for known lanes, whereas \fullfgrref{fig:bayesian-rmse-unknown} depicts the results for unknown lanes.
Each plot corresponds to a different incident scenario and illustrates RMSE for Bayesian $M_{abnormal}$, fresh $M_{abnormal}$, and $M_{ordinary}$, which does not utilize incident simulations and thus appears fixed.
We limit the plots to $j=1..5$ training simulations, because both $M_{abnormal}$ models typically perform closely for $j \geq 6$.

We see that compared to fresh $M_{abnormal}$, Bayesian $M_{abnormal}$ mostly obtains lower average RMSE and lower standard deviation for $j \leq 4$ simulations.
Hence by transferring knowledge from $M_{ordinary}$, Bayesian $M_{abnormal}$ can take better advantage of a low number of training simulations than can fresh $M_{abnormal}$, which has no such knowledge.

However, there are a few occasions where modeling afresh is more advantageous, e.g., scenarios ``L,S,B,B'' in \fgrref{fig:bayesian-rmse-known} and ``L,S,T,T'' in \fgrref{fig:bayesian-rmse-unknown}.
In addition, when QTIP generates exceedingly few training simulations ($j \leq 2$), $M_{ordinary}$ often outperforms both $M_{abnormal}$ models.
In conclusion, transfer learning can be advantageous in situations where QTIP resources are reasonably limited, e.g., when only a few computational nodes are available for parallel execution of what-if simulations.

\section{Summary of Key Points and Future Work} \label{sec:discussion}

We have presented the QTIP framework for real-time model adaptation under non-recurrent traffic disruptions.
QTIP is motivated by both a problem and an opportunity. 
The problem is that traffic prediction models must be adapted in real-time to properly deal with abnormal road conditions, yet current solutions fall short of addressing this need. 
The opportunity arises from In-Vehicle Monitoring Systems (IVMS), which provide immediate indication and information about incident occurrence.

Under incident conditions, QTIP generates the data required for model adaptation from real-time simulations of the affected road.
The simulations take advantage of IVMS as source of real-time incident information, while QTIP allows free choice of the complementary data-driven prediction models.
Our solution methodology thus combines two traditionally distinct approaches to problem modeling: "black box" machine learning algorithms on one hand, and ``white box'' transport engineering methods on the other hand.

\subsection{Findings and Implications} \label{sec:findings}

To evaluate QTIP, we have devised a proof-of-concept case study where incident conditions are represented as sudden road blocks on a major motorway in Denmark.
We have then experimented QTIP with several model types: Linear Regression, Gaussian Processes, and Deep Neural Networks.
Following are our main empirical findings.
\begin{enumerate}
    \item Our results verify measurably both the degradation in predictive quality if no model adaptation is performed, and the gain in predictive quality when QTIP is used for model adaptation (Section {\ref{sec:degradation}}).
    Additionally and as may be expected, the adapted prediction model improves when more incident parameters are known, regardless of model type (Section {\ref{sec:improvement}}).
    \item In most cases, the adapted model outperforms the non-adapted model, so that the mean relative RMSE improvement over all cases is $28.74$\% (Fig. {\ref{fig:rmse_improvement}}).
    There are also a few edge cases, where the non-adapted model performs better than QTIP, as we explain in Section {\ref{sec:improvement}}.
    \item As each simulation runs in under a minute on our single-PC platform, QTIP could yield an adapted model in real-time through parallel execution (Section {\ref{sec:realtime}}).
\end{enumerate}

Our findings thus imply that the long-standing problem of instantaneous model adaptation under incident occurrence is becoming more tractable, as In-Vehicle Monitoring Systems are increasingly deployed. 
This improvement in accuracy further implies real-world benefits for more effective incident management, and in particular,  our findings suggest considerable positive impact in circumstances of high traffic demand, during which efficient traffic management is most important.
Nevertheless, it is hard to assess the impact of this paper on traffic policy making, e.g., in terms of passenger time savings or the gains of quicker emergency response.
Rather, this is a proof-of-concept study with the objective of delivering a key theoretical message: that the suggested solution approach promotes the prospect of just-in-time prediction model adaptation.

\subsection{Limitations}

Every prediction framework has its limitations, and so does QTIP. For demonstrating the full potential of QTIP, this work examines the best case, namely $100\%$ availability of IVMS signals. Indeed, global trends suggest high adoption rate of IVMS in upcoming years. In addition, QTIP currently depends on an external module for calibration of simulations. Nevertheless, roads which are particularly prone to incidents may be pre-calibrated offline. It should also be noted that IVMS signals alone cannot resolve an inherent difficulty of traffic prediction: that full observability of network state requires a number of road sensors much larger than is typically available.
As such, any solution based wholly on simulations may yield a sub-optimal adapted model, no matter how many simulations it employs.

\subsection{Future Work} \label{sec:future}

The above discussion raises several interesting directions for future work:
\begin{itemize}
    \item It is expected that the adoption of IVMS technology will happen gradually. Thus, the performance of QTIP may be measured under varying, lower rates of availability of such signals.  
    \item The current QTIP framework can be extended to other test cases, i.e. road types, road conditions, and larger networks, to test e.g. other demand conditions and effects of route choice. It can also be extended to other traffic disruption types. All of these extensions may ultimately yield different predictive performances in QTIP.
    \item In large-scale network scenarios, the computational performance of QTIP may worsen. Methods for parallel simulations, sub-network selection, or multi-scale frameworks can be tested within QTIP to achieve timely prediction improvements.
    \item The integration of QTIP with existing systems for real-time traffic management \citep{kong2013developing, dynamit2015} can be piloted, to assess performance in practical deployment conditions.
    \item The data-driven prediction models used in this paper can be further enhanced as described in Section {\ref{sec:compare_performance}}, and additional model types can be experimented with, e.g., ARIMA or Gradient Boosting \citep{zhang2015gradient}.
    \item Finally, we also point out that the integration of online calibration within QTIP may provide increased prediction power to the overall framework (\fgr \ref{fig:qtip_operation_diagram}). The past $15$ years have been rich in online calibration of meso- and microscopic simulators \citep{antoniou2007nonlinear, qin2004adaptive, prakash2018improving}, and further work is needed to evaluate real-time performance and additional benefits of QTIP with online calibration.
\end{itemize} 

\section*{Acknowledgment}

The research leading to these results has received funding from the People Programme (Marie Curie Actions) of the European Union’s Horizon 2020 research and innovation programme under the Marie Sklodowska-Curie Individual Fellowship H2020-MSCA-IF-2016, ID number 745673.

\printbibliography

\end{document}